\documentclass{article}

\PassOptionsToPackage{numbers, sort&compress}{natbib}

\usepackage[preprint]{neurips_2025}

\usepackage{xspace}

\newcommand{\F}{\mathcal{F}}
\newcommand{\MD}{\texttt{MIDAS}\xspace}
\newcommand{\LD}{\texttt{LIDAS}\xspace}
\newcommand{\LN}{\texttt{LN-Scaling}\xspace}

\newcommand{\argdot}{\,\cdot\,}

\newcommand{\gaussup}[1]{\lceil #1 \rceil}
\newcommand{\gaussdown}[1]{\lfloor #1 \rfloor}

\newcommand{\changed}[1]{#1}

\usepackage{comment}
\usepackage{hyperref}
\usepackage{amsmath,amsfonts,bm}
\usepackage{url}
\usepackage{graphicx}
\usepackage{subcaption}
\usepackage{booktabs}
\usepackage[dvipsnames]{xcolor}
\usepackage[most]{tcolorbox}
\usepackage[font=small,labelfont=bf]{caption}
\usepackage{wrapfig}
\usepackage{algorithm}
\usepackage{algpseudocode}
\usepackage{multirow}
\usepackage[capitalise]{cleveref}

\definecolor{LinkBlue}{HTML}{1A4FA3}
\hypersetup{
    colorlinks,
    linkcolor={LinkBlue},
    citecolor={LinkBlue},
    urlcolor={LinkBlue},
}

\newtcolorbox{hypoanswer}[1][]{
  enhanced, breakable, sharp corners,
  boxrule=0pt,
  colback=Gray!5!white, colframe=Gray!5!white,
  borderline west={4pt}{0pt}{Gray!75!black},
  width=\linewidth,          
  enlarge left by=0mm,       
  enlarge right by=0mm,
  title={#1}
}

\title{Do Depth-Grown Models Overcome the Curse of Depth?\\ An In-Depth Analysis}

\makeatletter
\renewcommand\thefootnote{\fnsymbol{footnote}}
\makeatother

\author{%
\makebox[\textwidth][c]{%
\textbf{Ferdinand Kapl}\textsuperscript{\mdseries 1,2}\footnotemark[1] \hspace{.5pt} \footnotemark[3]\quad
\textbf{Emmanouil Angelis}\textsuperscript{\mdseries 1,2}\footnotemark[1] \hspace{.5pt} \footnotemark[3]\quad
\textbf{Tobias H\"oppe}\textsuperscript{\mdseries 1,2}\footnotemark[1] \hspace{.5pt} \footnotemark[3]}\\[0.25em]
\makebox[\textwidth][c]{%
\textbf{Kaitlin Maile}\textsuperscript{\mdseries 3}\footnotemark[2]\quad
\textbf{Johannes von Oswald}\textsuperscript{\mdseries 3}\footnotemark[2]\quad
\textbf{Nino Scherrer}\textsuperscript{\mdseries 3}\footnotemark[2]\quad
\textbf{Stefan Bauer}\textsuperscript{\mdseries 1,2}\footnotemark[2]}\\[0.4em]
\makebox[\textwidth][c]{%
\textsuperscript{1}\,Technical University of Munich \quad
\textsuperscript{2}\,Helmholtz AI, Munich \quad
\textsuperscript{3}\,Google, Paradigms of Intelligence Team}%
}

\begin{document}

\maketitle
\footnotetext[1]{Equal contribution.}
\footnotetext[2]{Provided equal in-depth feedback and guidance.}
\footnotetext[3]{Correspondence:\{\texttt{ferdinand.kapl,emmanouil.angelis,tobias.hoeppe}\}\texttt{@tum.de}.}

\makeatletter
\renewcommand\thefootnote{\arabic{footnote}}
\makeatother

\begin{abstract} 
Gradually growing the depth of Transformers during training can not only reduce training cost but also lead to improved reasoning performance, as shown by \mbox{MIDAS}~\citep{saunshi2024inductive}. Thus far, however, a mechanistic understanding of these gains has been missing. In this work, we establish a connection to recent work showing that layers in the second half of non-grown, pre-layernorm Transformers contribute much less to the final output distribution than those in the first half---also known as the \textit{Curse of Depth} \citep{sun2025curse, csordas2025language}. Using depth-wise analyses, we demonstrate that growth via gradual middle stacking yields more effective utilization of model depth, alters the residual stream structure, and facilitates the formation of permutable computational blocks. In addition, we propose a lightweight modification of \mbox{MIDAS} that yields further improvements in downstream reasoning benchmarks. Overall, this work highlights how the gradual growth of model depth can lead to the formation of distinct computational circuits and overcome the limited depth utilization seen in standard non-grown models.

\end{abstract}
    
\section{Introduction}
The remarkable success of large language models (LLMs) has been accompanied by immense computational and energy demands. This trend of training larger and larger networks is correlated with the increasing depth of model architectures \citep{kaplan2020scaling, hoffmann2022training}. As Transformers \citep{vaswani2017attention} lack recurrence, their computational capacity is directly linked to their depth. Greater depth enables more complex computations and improves capabilities like reasoning, compositional generalization and goal reaching \citep{petty2023impact, lad2024remarkable, wang20251000}. However, this pursuit of greater scale uncovers a critical inefficiency, as training such models is extremely resource-intensive \citep{varoquaux2025hype}. 

A core issue of the current paradigm is the observation that not all layers contribute equally to the final model's performance \citep{yin2023outlier, gromov2024unreasonable, li2024mix,men2024shortgpt}. \citet{csordas2025language} and \citet{sun2025curse} demonstrate that deeper layers of modern pre-layer Transformers tend to be less effective than their earlier counterparts, with many layers in the second half of the model contributing minimally to the final output---also known as the \textit{Curse of Depth} \citep{sun2025curse}. This observation, which highlights a kind of over-parametrization, is supported by findings that various architectures are remarkably robust to perturbations like skipping layers without significant performance loss \citep{lad2024remarkable, yin2023outlier}. The Curse of Depth represents a major resource inefficiency in today's paradigm. As highlighted by \citet{csordas2025language}, addressing these limitations is a pressing need for the community to avoid wasting valuable resources and to develop more efficient architectures that can leverage deep layers effectively.

A promising solution lies in gradually grown architectures, which dynamically expand a model's depth or width during training. These novel training strategies, such as gradual stacking \citep{gong2019efficient, reddi2023efficient}, enable efficient training by using layers from a smaller model to initialize the next stage. Of particular interest is the \MD method \citep{saunshi2024inductive}, which gradually increases depth by inserting new layers into the middle of the model. \MD has been shown not only to speed up training but also to improve performance on reasoning-heavy benchmarks, suggesting that this growth procedure introduces a favourable inductive bias. However, a clear mechanistic understanding of these gains has so far been missing.

\begin{figure}[t]
    \begin{center}
        \includegraphics[width=\textwidth]{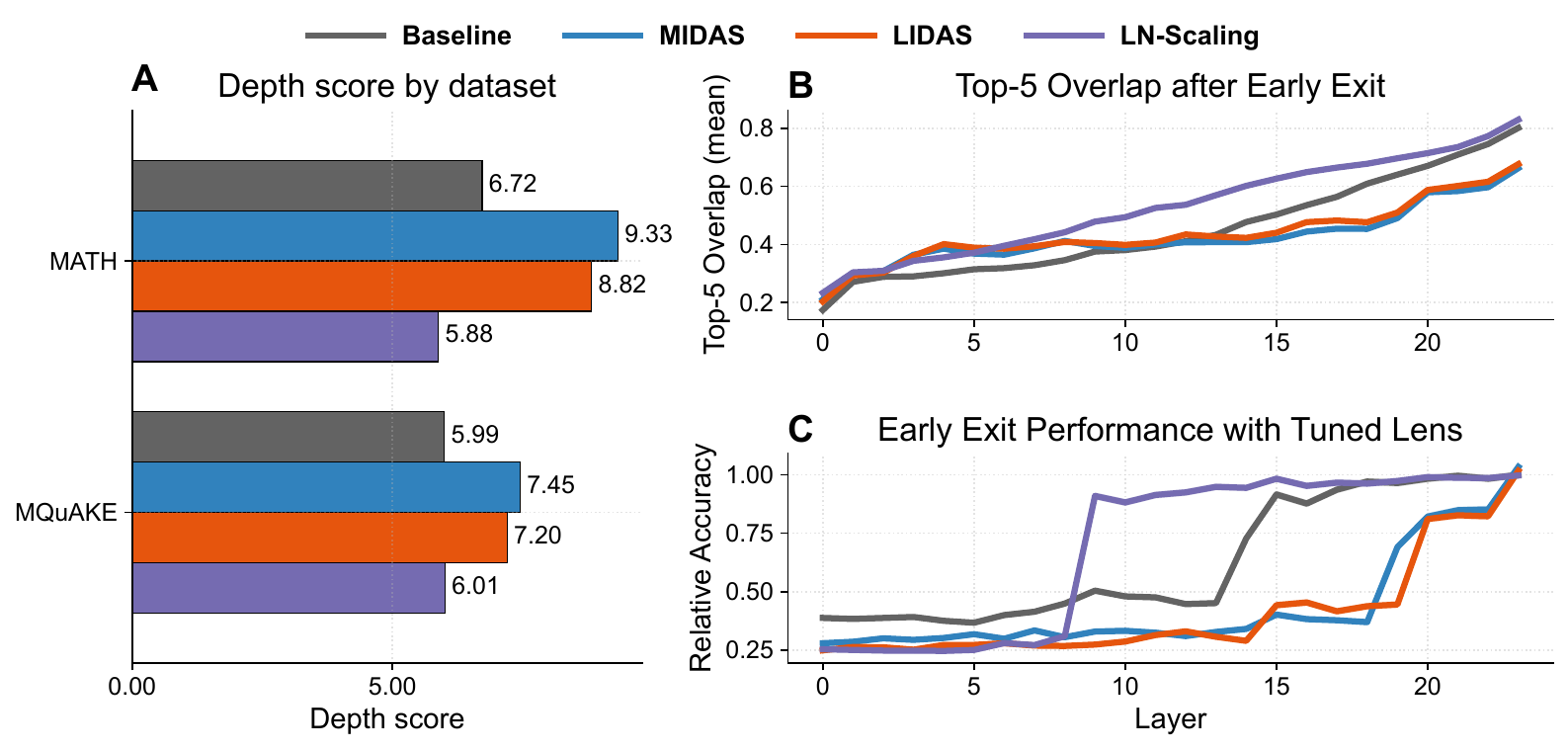}
    \end{center}
    \caption{ \textbf{Depth-grown models use their depth more (1.7B)}. (A) Depth score \citep{csordas2025language} on MATH \citep{hendrycks2021measuring} and MQuAKE \citep{zhong2023mquake}. Grown models (\MD, \LD) have consistently higher depth scores. (B) Top-5 overlap between each layer’s early-exit vocabulary and model’s final vocabulary on 20 prompts from GSM8K \citep{cobbe2021training}. Both grown models studied in this work exhibit lower overlap at later layers, indicating that these later layers still contribute additional features necessary for the final prediction. (C) Early-exit relative accuracy versus layer on \emph{Variable Assignment Math} reasoning primitive. The baseline reaches near its final performance early, whereas accuracy for \MD and \LD continues to rise up to the last layer. \changed{Using these metrics, however, \LN shows no discernible benefit over the baseline in depth utilisation.}}
    \label{fig:midas:depth_efficiency}
\end{figure}

\changed{In this work, we establish a direct connection between gradual depth growth and the ``Curse of Depth'' \citep{sun2025curse}, providing a mechanistic understanding of how gradual depth growth procedures can lead to more effective utilisation of a model’s depth by altering its computational structure. Using analysis tools from \cite{belrose2023eliciting} and \cite{csordas2025language}, we show that gradual stacking counteracts the patterns of diminishing returns observed in non-grown models and gives rise to qualitatively different depth-wise computation. We summarize our contributions below:}

\begin{itemize}
\item \textbf{\MD reproduction on different backbones.} We reproduce the core \MD results on SmolLM-v1 backbones (360M and 1.7B), trained with autoregressive next-token prediction, confirming that gradual depth growth improves reasoning performance over a conventionally trained, non-grown baseline with a $1.29\text{x}$ improvement in training speed.

\item \textbf{Novel gradual depth growth strategy \LD.} \changed{Based on the motivation of MIDAS, i.e., its connection to Looped Transformers and symmetric functional behaviour}, we propose \LD, an improved growing strategy that duplicates the layer-wise middle rather than the block-wise middle while preserving the inductive bias of growing. Across scales, \LD matches or exceeds \MD and conventionally trained models in reasoning benchmarks without degrading Negative Log-Likelihood (NLL) or knowledge performance. \changed{Additionally, \LD results in a more symmetric weight structure than \MD and aligns its attention sublayers in the middle of the network better with the residual stream.}

\item \textbf{\changed{Extensive analysis of improved and altered depth utilization in grown models.}} We provide an in-depth analysis of how gradual depth growth alters computation and representation in LLMs, providing mechanistic insights into these findings. We show that grown models utilize their depth more efficiently than conventionally trained baselines, \changed{by reaching their final performance only at the very last layer}. Furthermore, we demonstrate that grown models develop permutable computational blocks in the middle of the network, with each layer within a block fulfilling a specific cyclical role.
\end{itemize}

Overall, this work provides a first mechanistic understanding of how gradual depth growth can counteract the Curse of Depth, potentially leading to more efficient and capable language models.

\section{Related Work}

\textbf{Growing Neural Networks.} Early on, researchers recognized the advantages of training neural networks one layer at a time \citep{hinton2006fast, bengio2006greedy} to overcome the challenges of learning long-term dependencies with gradient descent \citep{bengio1994learning}. More recently, this concept has been re-explored for large language models (LLMs) through gradual stacking \citep{gong2019efficient, reddi2023efficient, du2024stacking} and depth up-scaling \citep{kim2023solar}. Alternative growing strategies include masked structural growth \citep{yao2023masked}, function-preserving expansions \citep{gesmundo2023composable} and learned linear growth operators \citep{wang2023learning}. 

\textbf{Adaptive Architectures.} A potentially complementary strategy to growing is using adaptive architectures that dynamically adjust their computational graph or parameters based on their input data by using a larger, pre-trained network more efficiently, including mixture of experts approaches \citep{jacobs1991adaptive, shazeer2017outrageously, csordas2024moeut} or early exiting \citep{teerapittayanon2016branchynet,xin2020deebert}. 
Recent approaches apply adaptive token-level computations \citep{bae2025mixture}, nested models for elastic inference \citep{devvrit2024matformer} or depth-wise looping \citep{giannou2023looped,yang2023looped,vonoswald2025mesanetsequencemodelinglocally}. 

\textbf{Depth of Neural Network Architectures.} While depth is a key factor correlated with network performance \citep{csordas2025language}, recent research has found that deeper layers in LLMs are often redundant and less effective. \citet{sun2025curse} have termed this phenomenon the Curse of Depth, which suggests deeper layers contribute minimally to learning. \changed{To counteract, they propose \LN, which scales each layer’s LayerNorm activation to suppress depth-induced variance growth so deeper layers learn usefully. Complementary to such architectural and normalisation changes, \citet{dey2025don} introduce CompleteP, a depth-and width-aware parameterisation that achieves depth-wise hyperparameter transfer, enabling compute-efficient training of very deep Transformers across a broad range of width–depth aspect ratios.} Studies on models like GPT-2 show that their middle and deep layers exhibit remarkable robustness to significant perturbations, including layer swapping and deletion \citep{yin2023outlier,lad2024remarkable}. This over-provisioning has inspired various layer intervention strategies, such as skipping, swapping, or parallelization, to improve efficiency \citep{lad2024remarkable, sun2025transformer}.

\textbf{Reasoning.} For solving challenging tasks, recent work has shifted focus to recurrence and looping \citep{geiping2025scaling, saunshi2025reasoning} to improve model reasoning and leverage depth scaling for enhanced internal "thinking" \citep{chen2025inner}. These methods scale up test-time computation to allow models to iteratively refine their answers. Complementary to these approaches, our work focuses on identifying and leveraging computational blocks within depth-grown neural networks to improve reasoning, rather than relying on a fixed, recurrent process.

\section{Two Depth-Grown Transformers: MIDAS \& LIDAS}
In this section, we first formalise the growth operator on a fixed architecture class $\F$ and recover \MD \citep{saunshi2024inductive} as a special case. We then introduce \LD, which inserts a new middle block constructed by interleaving its neighbours to provide a stronger initialisation. Finally, using models from the SmolLM-v1 family \citep{benallal2024smollmcorpus}, we present empirical results on aggregated reasoning and knowledge benchmarks, showing that both gradual-depth growing methods outperform a conventionally trained, non-grown baseline and \changed{LayerNorm-Scaling (\LN \citet{sun2025curse})} on reasoning tasks while remaining on par with general language-modelling performance.

\subsection{The Growing operator}
\label{sec:growing_operator}
We fix a base architecture class $\F$ (width, heads, embedding size, etc. are fixed) and vary only depth.
Let $f_L \in \F$ denote a model with $L$ Transformer layers, written as an ordered list $f_L = [\ell_0,\dots,\ell_{L-1}]$.
A (depth) \emph{growth operator} $G : \F \times \mathbb{N} \to \F$ maps an $L$-layer model to an $(L+b)$-layer model, such that $G(f_L; b) = f_{L+b}$, where $b \in \mathbb{N}$ is the \emph{block size} (the number of layers added per growth step). Following \citet{saunshi2024inductive}, we consider growth operators that insert new layers in the centre of the model and keep the block size $b$ fixed across growing stages.

\begin{figure}[htb]
\begin{center}
\includegraphics[width=0.8\textwidth,trim={5mm 3mm 5mm 5mm},clip]{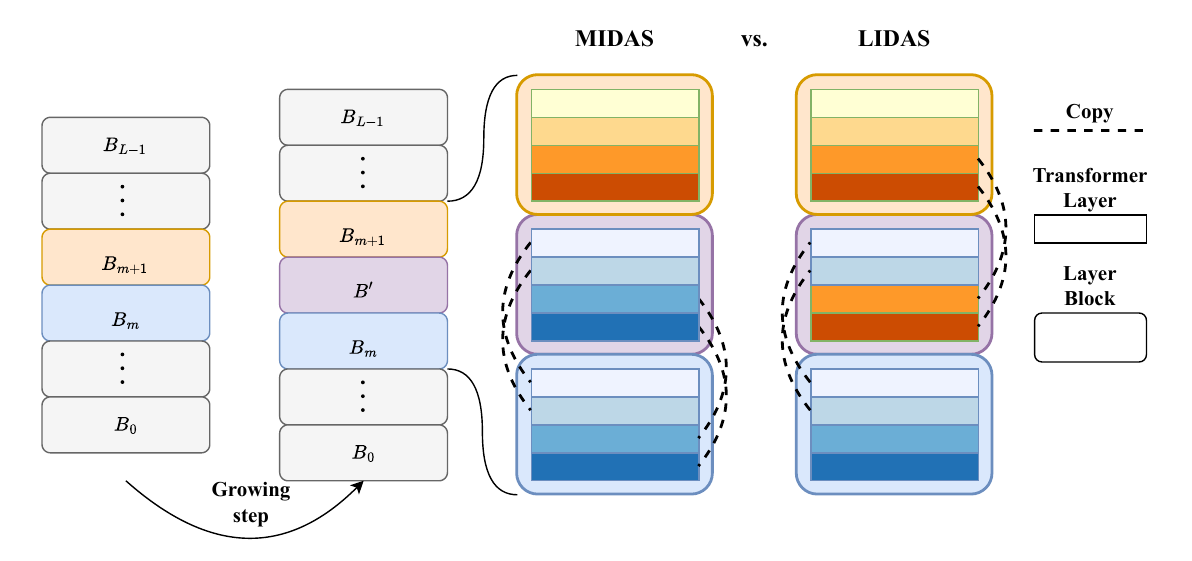}
\end{center}
\caption{ \textbf{Illustration of growing strategies with block size 4}: \MD vs. \LD, with an even number of existing blocks. \MD \citep{saunshi2024inductive} simply copies $B^{\prime}=B_m$, which is the block preceding mid-depth. When seen from a block-wise perspective instead of a layer-wise perspective, our proposed variant \LD may be interpreted as forming $B^{\prime}$ from the two blocks surrounding the mid-depth \changed{by combining the first two layers of $B_{m+1}$ with the last two layers of $B_m$}. This small difference in initialization leads to significantly improved performance as shown in \cref{tab:exp:results}.}
\label{fig:midas:sketch}
\end{figure}

The following strategies use layer duplication to initialize new layers within the newly inserted block. This consists of deep copying all parameters within the layer, including their optimizer state. The result is two initially identical copies at different depths of the model, thus allowed to diverge as training continues.

\textbf{\MD.} When depth increases by $b$ layers per stage, at stage $n$ we have $L = n b$ and can partition $f_L$ into $n$ contiguous blocks of size $b:$
\begin{equation}
    f_L = [B_0 \,\|\, B_1 \,\|\, \cdots \,\|\, B_{n-1}], \quad B_j = [\ell_{jb},\dots,\ell_{(j+1)b-1}].
\end{equation}
 Let $m_b= \gaussup{\frac{n}{2}}-1$ denote the \emph{middle block} index. Middle gradual stacking inserts a new block $B^\prime$ immediately after $B_{m_b}$, i.e. 
\begin{equation}
    G(f_L; b) = [B_0 \,\|\, \cdots \,\|\, B_{m_b} \,\|\, B^{\prime}   \,\|\, B_{m_b+1} \,\|\, \cdots \,\|\, B_{n-1}].
\end{equation}
 If $B^{\prime} = B_{m_b}$ (copying the middle block), we recover \MD as proposed in \citet{saunshi2024inductive}. 

\textbf{\LD.} Since we are constrained by the block patterning in \MD, we propose Layer-wise mIDdle grAdual Stacking, or \LD, in which we consider the \emph{middle layer} $m_l = \gaussup{\frac{L}{2}}$ to be the central point of the growing operation. We then construct a new block $B^{\prime} = [l_{m_l - \gaussup{b/2}}, \dots, l_{m_l + \gaussdown{b/2}}]$, around the middle layer $l_{m_l}$, which is inserted after the layer $l_{m_l + \gaussdown{b/2}}$. For an odd number of blocks, \MD and \LD coincide by selecting the same layers. They differ for an even number of blocks, shown from a block-wise perspective in \cref{fig:midas:sketch}. Further details can be found in \cref{sec:apx:smollm}.

\textbf{Training runs and schedules.}
A training run is specified by (i) the model class $\F$, (ii) the target depth $L_{\mathrm{final}}$, (iii) the initial depth $L_0$ (typically $L_0=b$), (iv) a fixed block size $b$, and (v) a stage schedule $\{T_s\}_{s=0}^{S-1}$ (training steps per stage). Starting from $f_{L_0}$, after each stage $s$, we apply $G(\argdot; b)$ to obtain $f_{L_{s+1}}$ with  depth $L_{s+1}=L_s+b$.
We repeat until $L_{S-1} = L_{\mathrm{final}}$.

\subsection{Experiments}
\label{sec:exp_benchmarks}
\textbf{Setup.} We evaluate and compare the two growing methods, \MD and \LD, against a conventionally non-grown baseline \changed{and one alternative method \LN}. We use the 360M and 1.7B models from the SmolLM-v1 family \citep{benallal2024smollmcorpus}, \changed{trained on 200B and 400B tokens respectively}, to probe scaling behaviour. All models are trained from scratch on the SmolLM-Corpus, a curated mixture of educational and synthetic texts as well as mathematics and code. Due to their favourable efficiency–performance trade-off, these models enable competitive evaluation within a constrained computational budget. For all grown models we present, we use the block size $b=4$ and a PROP-1 growing schedule (see \cref{sec:apx:smollm} for details).

\begin{figure}[thp!]
    \begin{minipage}[b]{1.0\textwidth}
     \resizebox{0.9999\textwidth}{!}{ 
        \addtolength{\tabcolsep}{-0.2em}
        \begin{tabular}{ll|c|cc|cc|cc|cc}
        \toprule
        \multicolumn{2}{c|}{} &\multicolumn{7}{c|}{Standard cooldown} &\multicolumn{2}{c}{Math cooldown}  \\ \cmidrule(){3-11}
        \multicolumn{2}{c|}{\bf } &\multicolumn{1}{c|}{\bf} &\multicolumn{1}{c}{\bf Open-book} &\multicolumn{1}{c|}{\bf Closed-book}&\multicolumn{1}{c}{\bf} &\multicolumn{1}{c|}{\bf} &\multicolumn{2}{c|}{} &\multicolumn{2}{c}{}  \\
        \multicolumn{2}{c|}{\bf } &\multicolumn{1}{c|}{\bf Holdout Set} &\multicolumn{1}{c}{\bf Q\&A} &\multicolumn{1}{c|}{\bf Q\&A}&\multicolumn{1}{c}{\bf Lambada} &\multicolumn{1}{c|}{\bf Hellaswag } &\multicolumn{1}{c}{\bf Math Word} &\multicolumn{1}{c|}{\bf Primitives } &\multicolumn{1}{c}{\bf Math Word} &\multicolumn{1}{c}{\bf Primitives } \\
        \multicolumn{2}{c|}{\bf } &\multicolumn{1}{c|}{(NLL $\downarrow$)} &\multicolumn{1}{c}{(F1 $\uparrow$)} &\multicolumn{1}{c|}{(F1 $\uparrow$)} &\multicolumn{1}{c}{(Acc $\uparrow$)} &\multicolumn{1}{c|}{(Acc $\uparrow$)} &\multicolumn{1}{c}{(Acc $\uparrow$)} &\multicolumn{1}{c|}{(Acc $\uparrow$)} &\multicolumn{1}{c}{(Acc $\uparrow$)} &\multicolumn{1}{c}{(Acc $\uparrow$)} \\
        \toprule
        \parbox[t]{2mm}{\multirow{4}{*}{\rotatebox[origin=c]{90}{360M}}} & Baseline &2.18 &22.89 &14.50 &43.35 & 39.97 &3.69 &30.06 & 8.10 & 33.12  \\
        \cmidrule(){2-11}
        & \changed{\LN}   &\textbf{2.16} &23.14 &\textbf{14.89} &42.17 &40.0 &2.89 &\textbf{31.38} &8.45 &41.26 \\
        &\MD          &2.18 &24.57 &13.75 &43.31 &40.36 &\textbf{4.39} &28.18 &\textbf{13.43} & 35.14\\
        & \LD   &\textbf{2.16} &\textbf{26.63} &14.57 &\textbf{44.03} &\textbf{40.58} &4.36 &31.20 &12.30 &\textbf{50.36} \\
        
        \toprule
        \parbox[t]{2mm}{\multirow{4}{*}{\rotatebox[origin=c]{90}{1.7B}}} & Baseline &\textbf{1.96} &29.57 &18.61 &50.05 &46.28 &13.75  &34.84  & 23.28  & 42.77   \\
        \cmidrule(){2-11}
        & \changed{\LN}       &1.97 &29.11 &18.63 &48.94 &45.44 &11.0  &44.38 & 17.84  &50.58   \\
        & \MD         &1.97 &28.80 &18.50 &50.81 &46.19 &16.07  &40.88 & 24.01  & \textbf{55.46}  \\
        & \LD       &\textbf{1.96} &\textbf{29.84} &\textbf{19.08} &\textbf{51.41} &\textbf{46.32} &\textbf{18.59}  &\textbf{47.34} & \textbf{24.60}  &53.00   \\
        
        \bottomrule
        \end{tabular}
        }
        \small \captionof{table}{\textbf{Performance comparison of a standard transformer baseline, LayerNorm-Scaling, and the two grown models \MD and \LD}. We reproduce the findings of \citet{saunshi2024inductive} and observe that grown models match the baseline in training objective (NLL), standard Q\&A benchmarks as well as Lambada. Grown models, especially \LD, outperform the non-grown baseline on reasoning-heavy tasks such as Math Word and Primitives. \changed{\LN on the other hand, achieves only minor improvements, which diminish when scaling to the larger model.}}
        \label{tab:exp:results}
    \end{minipage}
\end{figure}

\textbf{Benchmarks.} We report negative log-likelihood (NLL) on a held-out validation set from the SmolLM-Corpus. We follow the knowledge and reasoning benchmarking suite reported in \citet{saunshi2024inductive}. The knowledge-based benchmarks are split into Open-book Q\&A with provided context (TyDiQA-GoldP, SQuADv2, DROP, QuAC, CoQA), and Closed-book Q\&A without context (TriviaQA, TyDiQA-NoContext, NaturalQuestions, WebQuestions), evaluated zero-shot. We additionally add Lambada \citep{paperno2016lambada} and HellaSwag \citep{zellers2019hellaswag} in their classical settings. For reasoning, we report the aggregated performance on Math Word problems (SVAMP \citep{patel2021nlp}, ASDiv \citep{miao2021diverse}, and MAWPS \citep{koncel2016mawps}) and reasoning primitives, which are a suite of synthetic tasks designed by \citet{saunshi2024inductive} to specifically investigate reasoning performance on a smaller scale, both evaluated under five-shot prompting as done previously. For the exact score breakdown, we refer to \Cref{sec:apx:detailed_results}.

\looseness-1 \textbf{Results.} Aggregated results are shown in \cref{tab:exp:results}. Consistent with the findings of \citet{saunshi2024inductive}, we observe that depth-grown models (both \MD and \LD) outperform the baseline on reasoning-heavy tasks (i.e.,\ Math Word and Reasoning Primitives). On the remaining benchmarks (Open-book Q\&A, Closed-book Q\&A, and Lambada), we observe little deviation from the baseline model with \LD being slightly superior to \MD. In addition, we observe a 29\% speedup in training for depth-grown models over a conventionally trained, non-grown baseline (\cref{tab:apx:flops}). \changed{\LN, on the other hand, largely tracks the baseline on NLL and knowledge tasks, offering only small gains at 360M that disappear at 1.7B, while trailing the grown variants on reasoning}. In summary, our results reproduce the observation of \citet{saunshi2024inductive}. \MD outperforms a conventionally trained baseline on reasoning-heavy tasks, and our proposed method \LD further strengthens this effect without degrading NLL at 1.7B. To stabilise results on Math Word, we additionally report the performance of models which are finetuned on the OpenWebMath dataset. While the relative order stays the same, we observe for the 360M model that the improvements for the grown models become more pronounced. Additionally, we report the performance of the 360M baseline and its growing variants on another seed in \cref{tab:app:results_add_seed}, confirming that the above findings are robust. However, the reasons behind these gains remain unclear. Therefore, we turn next to a detailed analysis of the 1.7B models, aiming to characterize how \MD and \LD may mechanistically differ from the baseline and how this could lead to improved performance in reasoning tasks.

\section{Depth Analysis}
\label{sec:depth_analysis}
Motivated by the confirmed observations that gradually depth-grown Transformers seem to yield increased reasoning abilities, we now investigate how gradual depth growth reshapes computation across depth in the trained models. To this end, we first examine early-exiting performance for every layer with TunedLens \citep{belrose2023eliciting} to test how much performance degrades when we exit early. Next, we run various interventions on the models, such as swapping contiguous blocks of layers, to test whether grown models form permutable circuits, and how sensitive each method is to late-layer ablations. We then analyse the layer-wise roles within blocks by measuring the similarity between each layer's contribution and the residual stream. Finally, we compare \MD with \LD on weight symmetry and contribution per attention matrix, connecting it to benchmark results of the previous section. All analyses are conducted on the 1.7B variant described in \cref{tab:exp:results}, and analogous results for the 360M models are reported in \cref{sec:supp_small_models}. A detailed description of the setups for each analysis can be found in \cref{sec:apx:exp_setup}. For notation, we follow \cite{csordas2025language}: $h_{i+1}$ denotes the residual stream after transformer layer $l_i$, $a_i$ the layer's attention output and $m_i$ the output of the MLP.

\subsection{Does Depth Growth Lead to Different Depth Utilization?}
\label{sec:depth_utilization}
\begin{hypoanswer}[]
\textbf{Hypothesis.} Gradual depth-grown Transformers (with \MD and \LD) utilize model depth more efficiently than conventionally trained, non-grown Transformer baselines.

\medskip
\textbf{Evidence.} Skipping late layers degrades prediction accuracy substantially more for \MD and \LD than for the baseline, which coincides with an increased depth score.
\end{hypoanswer}

\textbf{Experiments.}
To investigate the contribution of deeper layers, we evaluate intermediate representations via a Tuned Lens \citep{belrose2023eliciting}. Concretely, for each layer $l_i$, we train a small affine adapter on a split of FineWeb-Edu \citep{penedo2024fineweb} that maps that layer’s residual output to the hidden representation consumed by the final normalization; we then obtain logits by applying the model’s final normalization and unembedding \citep{belrose2023eliciting}, enabling early-exit at every depth\footnote{Note that this should result in more accurate predictions than naively applying the unembedding matrix at every layer (LogitLens \citep{logitlens}), as done in \citet{csordas2025language}.}. Subsequently, we quantify depth utilization by the top-5 vocabulary overlap of their predicted vocabularies (\cref{fig:midas:depth_efficiency}B), and early-exiting accuracy on a reasoning primitive (\cref{fig:midas:depth_efficiency}C). Finally, we compute the depth score \citep{csordas2025language} to summarize where computation occurs along the network by estimating each layer’s influence on future tokens (\cref{fig:midas:depth_efficiency}A). For further details, we refer to \cref{sec:apx:exp_setup}, and for results on the 360M model, to \cref{sec:apx:360_results}.

\textbf{Interpretation.} For \MD and \LD, \cref{fig:midas:depth_efficiency}B shows that early-exit predictions differ substantially more from the final logits than in the baseline (lower top-5 overlap), indicating that later layers in the grown models add features to the residual stream that are required for the final prediction. In \cref{fig:midas:depth_efficiency}C, the baseline reaches its final performance by Layer 18, whereas accuracy for both grown models continues to improve up to the last layer. Lastly, \cref{fig:midas:depth_efficiency}A consistently reports higher depth scores for the grown models across datasets, most notably on math tasks, indicating that more computation is concentrated in later layers. \changed{Interestingly, \LN, designed by \citet{sun2025curse} to reduce output variance in deeper layers of pre-layernorm transformers and thereby improve the utilization of later layers, exhibits the opposite behavior here. Concretely, \LN shows a lower depth score, stronger overlap with final tokens in earlier layers, and achieves peak performance earlier than both the baseline and grown models for the 1.7B (\cref{fig:midas:depth_efficiency}) and 360M (\cref{fig:midas:depth_efficiency_360}) models.}

\subsection{Does Depth Growth Form Permutable Computational Blocks?}
\label{sec:comp_blocks}
\begin{hypoanswer}[]
\textbf{Hypothesis.} Non-grown models depend on their specific layer ordering. Depth-grown models, on the other hand, develop computational blocks that are robust to block-level ordering interventions.

\medskip
\textbf{Evidence.} Reduced performance degradation under multi-layer perturbations indicates lower layer order dependence and greater robustness of \MD and \LD.
\end{hypoanswer}

\textbf{Experiments.} To evaluate layer functional independence, we swap contiguous sub-blocks of sizes $\{1, 2, 4, 8\}$ and measure the effect on downstream performance. While these experiments can indicate robustness, we can also observe how commutative sub-blocks are, as the local order of layers is preserved when swapping larger blocks. 

\begin{figure}[htb]
\centering
    \centering
    \includegraphics[width=\linewidth]{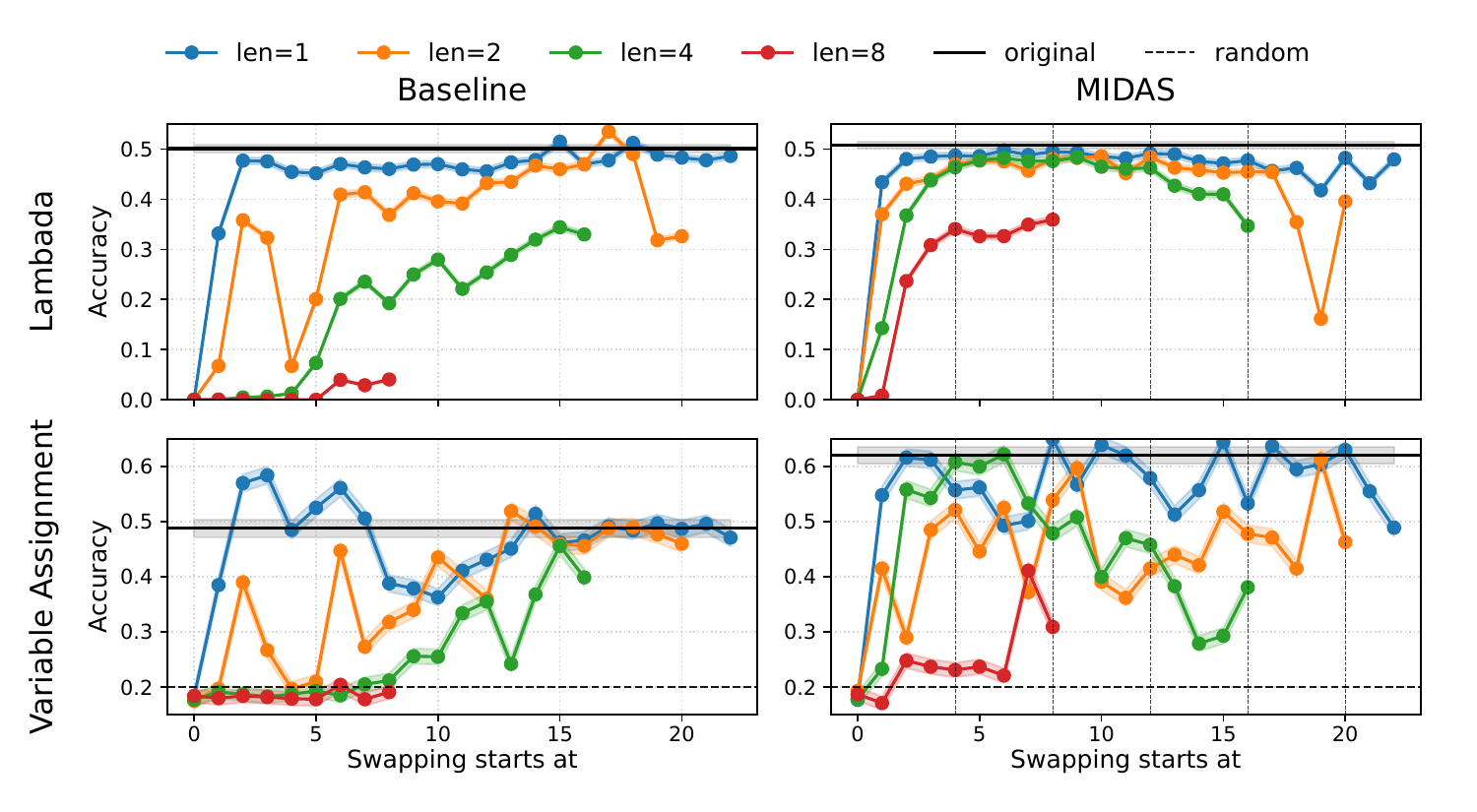}
    \caption{ \textbf{Effect of swapping blocks of layers on Lambada (top row) and the reasoning primitive \emph{Variable Assignment Math} (bottom row).} \MD is more robust to interventions for larger blocks in the middle of the network: the degradation in performance for \MD is much smaller for swapping blocks of larger sizes $\{2, 4, 8\}$ compared to the baseline, especially for Lambada. In Appendix \cref{fig:apx:big_swap}, we present results including \LD.}
    \label{fig:main_skipping_swapping_lambada}
\end{figure}

\textbf{Interpretation.} Swapping just single layers does not affect the performance of the baseline and grown models much (\cref{fig:main_skipping_swapping_lambada}), except for the input layers. This observation aligns with findings from \citet{lad2024remarkable}. If we increase the number of consecutive layers that we swap, the accuracy of the baseline quickly starts to deteriorate. In contrast, grown models allow swapping blocks of up to size four with a relatively small decrease in performance, and we observe even less performance degradation when swapping blocks, indicating less order dependence of these blocks. The grown models even reach non-random performance when swapping middle 8-layer blocks compared to the baseline, whose performance drops to random. In general, the degradation is lower on the language-modelling task (\cref{fig:main_skipping_swapping_lambada} top row) compared to the reasoning primitive (bottom row). Taken together, these effects are most consistent with the emergence of computational blocks whose internal order matters less than the presence of the block as a unit, matching the qualitative behaviour in \cref{fig:main_skipping_swapping_lambada}.

\subsection{Does gradual growth form layer-wise patterns?}
\label{sec:growth_patterns}
\begin{hypoanswer}[]

\textbf{Hypothesis.} The block-wise growing introduces a cyclical pattern in the architecture such that each layer within a block fulfils a certain role.

\medskip
\textbf{Evidence.} The contribution of the attention sublayer, in norm and cosine similarity, repeats in each block. When performing causal interventions, the effect for each layer within a block also repeats. Reversing the order of layers within and especially across blocks destroys the performance of grown models more than swapping, where local order is more preserved.
\end{hypoanswer}

\textbf{Experiments.} Using the tools of \citet{csordas2025language}, we compute for each attention layer its cosine similarity to the residual stream ($\frac{a_i \cdot h_i}{||a_i||||h_i||}$) and its mean relative contribution $\frac{||a_i||}{||h_i||}$. We then intervene by skipping a transformer layer or sublayer and track the relative changes in downstream computations under two regimes: (i) propagated: zeroing that component’s contribution to the residual stream and forwarding this change to all downstream layers; and (ii) local: removing a layer’s contribution from all subsequent inputs separately to isolate pairwise source--target dependencies. Finally, we assess the effect of reversing the order of four consecutive layers and comparing the outcome to results from \cref{fig:main_skipping_swapping_lambada}. A detailed explanation of the interventions can be found in \cref{par:apx:intervention_protocol}.

\begin{figure}[htb]
    \begin{center}
        \includegraphics[width=\textwidth]{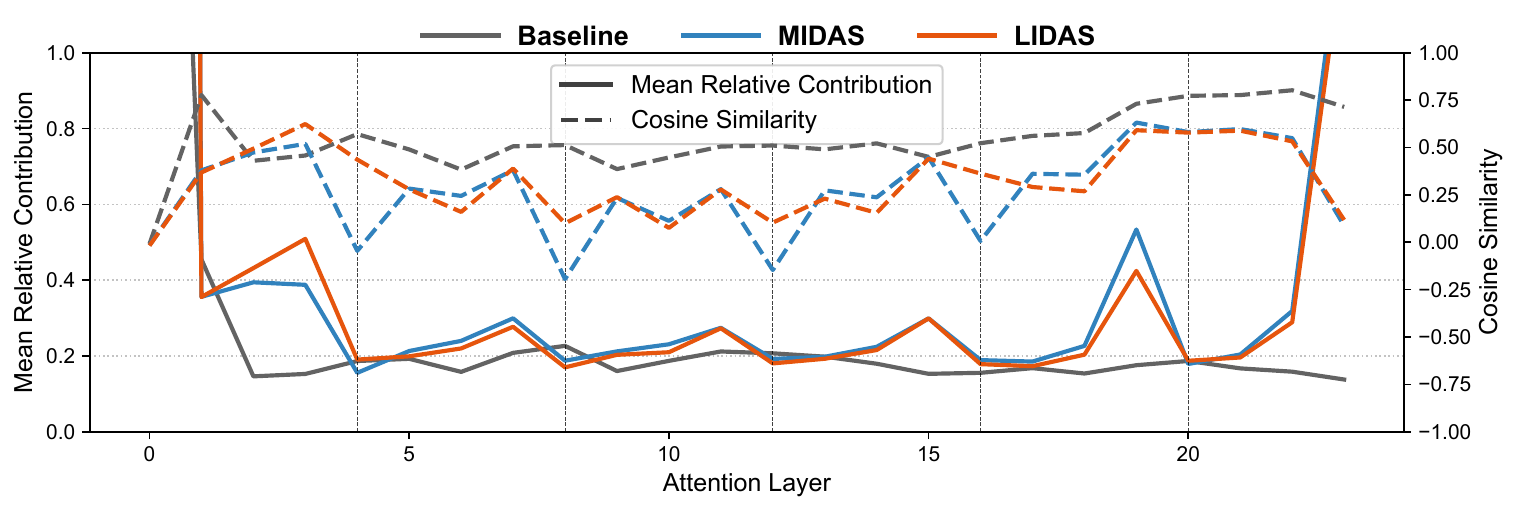}
    \end{center}
\caption{ \textbf{Attention layer contributions to the residual stream.} Grown models exhibit a highly cyclical pattern in the centre of the network. The mean relative contribution of the attention sublayer to the residual stream increases throughout a block (whose first layer is denoted by a vertical line) and has its largest contribution in the last layer of each block. While the cosine similarity between the output of each attention layer and the residual stream is relatively flat for the baseline, the pattern for the grown models again depends on the block size and the relative position of the layers within each block. Notably in \MD, the first attention sublayer of a block has a very low cosine similarity to the residual stream, while for \LD the attention contributions align more with the residual stream.}
\label{fig:main_cyclical_attention}
\end{figure}

\textbf{Interpretation.} Grown models exhibit a highly cyclical pattern in the middle, where the effect is especially visible for the attention sublayer (\cref{fig:main_cyclical_attention}). The mean relative contribution of the attention sublayer always grows from its lowest point at the first layer of every block to its highest point at the last layer of the block. The highest spike across depth is always at the final layer of the last block in the middle of the network, i.e., the overall second-to-last block.
For \MD the cosine similarity of the attention sublayers in the middle, similarly to their contributions, always rises from around zero, adding orthogonal features, or slightly negative, weakening or erasing features, to the highest but only slightly positive cosine similarity at the end of each block. The pattern for \LD is a little bit less clear, but the cosine similarity never drops as low as \MD, potentially adding features from subspaces that are better aligned with the residual stream across the whole block. 

\begin{figure}[htb]
\centering
    \begin{subfigure}[b]{0.5\linewidth}
    \centering
    \includegraphics[width=\linewidth]{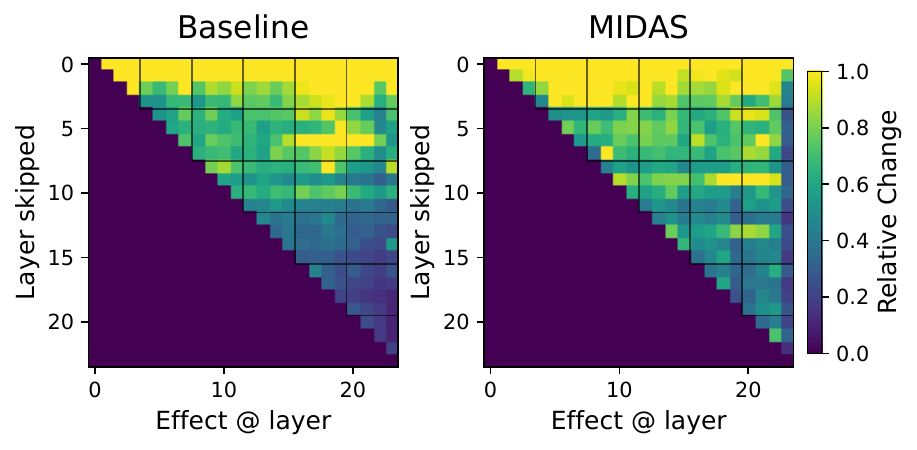}
    \caption{\centering Propagated Effect}
    \end{subfigure}%
    \begin{subfigure}[b]{0.5\linewidth}
    \centering
    \includegraphics[width=\linewidth]{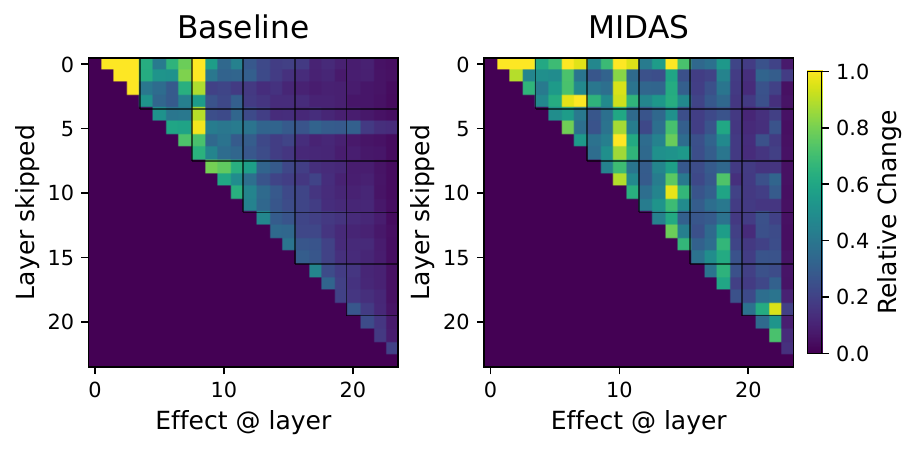}
    \caption{\centering Local Effect
    }
    \end{subfigure}
\caption{ \textbf{Baseline vs. \MD. Effect of skipping a layer on downstream layer contributions for \emph{future} tokens}. (a) \MD relies more on later layers than the baseline for future computations. Especially skipping the second layer of each mid-block strongly impacts the immediately following layer. (b) For \MD, the third layer of every block in the middle directly depends on all previous computations. We refer to \cref{fig:apx:big_future_effects} and \cref{fig:apx:big_future_local_effects} in the Appendix for results including \LD.}
\label{fig:main_midas_heatmaps}
\end{figure}

Turning towards interventions, by skipping a layer, the most pronounced disruption to future computations arises when skipping the second layer of each block (aside from the earliest layers), with often the biggest observed relative change in the immediate layer after it, i.e., in each block’s third layer (\cref{fig:main_midas_heatmaps}a). We hypothesize that the second layer prepares features for future computations. If we measure the relative change on the following layers directly, we notice a clear and striking pattern (\cref{fig:main_midas_heatmaps}b). For future computations, the third layer of every block directly depends on the features of almost all previous layers, potentially performing an aggregating operation. The direct change of removing the output of a previous layer is less on deeper blocks that can depend on more inputs simultaneously, i.e., visually a fading pattern. The last block mostly depends on the final aggregation and strengthening of relevant features performed by the second-to-last block.

\begin{figure}[htb]
\centering
    \centering
    \includegraphics[width=\linewidth]{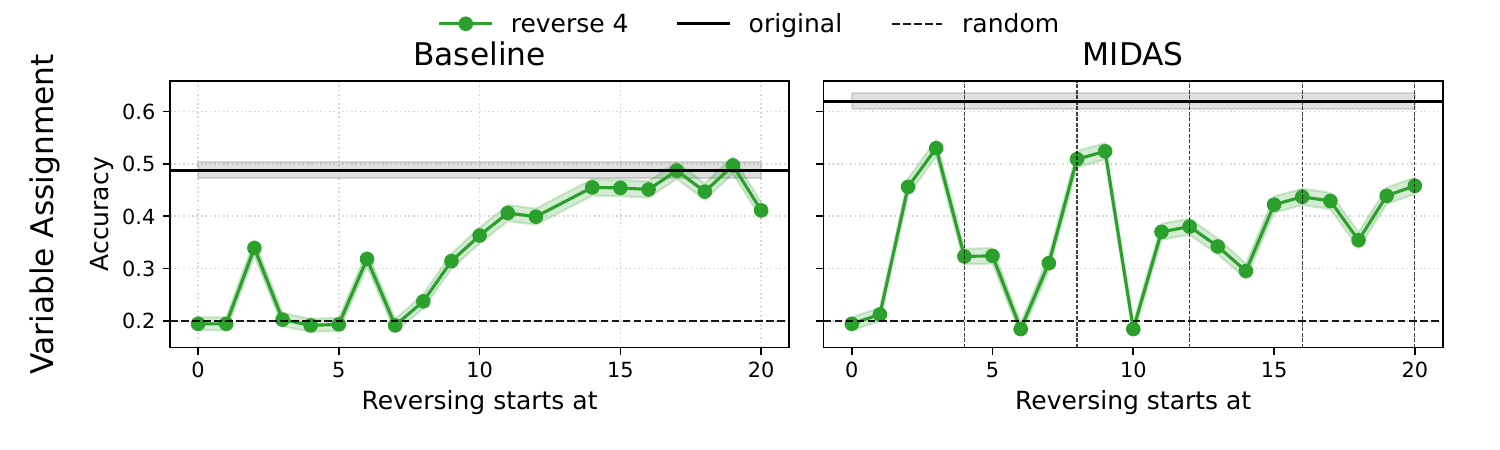}
    \caption{ \textbf{Effect of reversing the order of four consecutive layers on reasoning primitive.} Reversing the order of layers within a block (first layer of each grown block as vertical grey lines; right figure) of size 4 degrades the performance for grown models more than swapping the same number of layers ($\text{len}\!=\!2$ in \cref{fig:main_skipping_swapping_lambada}). The baseline is more robust to reversing the order of the later layers, while \MD is especially sensitive to reversing the order across grown blocks, i.e., the last two and first two layers of consecutive blocks. Starting to reverse at these positions, which correspond to layer index 6, 10, and 14, always results in a substantial drop in performance. Appendix \cref{fig:apx:big_reverse} shows results including \LD and an additional dataset.}
    \label{fig:main_reverse_prim}
\end{figure}

Reversing the order of four consecutive layers (\cref{fig:main_reverse_prim}) reduces performance in the grown model far more than swapping pairs of two or four layers ($\text{len}\!=\!2,4$ in \cref{fig:main_skipping_swapping_lambada}), where local order is more preserved. The baseline is comparatively robust to reversals involving later layers, which aligns with the hypothesis from \cite{csordas2025language} that later layers in pre-layernorm transformers refine the current output distribution with less order dependence. By contrast, the grown model is most brittle when the reversal straddles block boundaries (last two layers and first two layers of consecutive blocks), showcasing that the order of layers within a block matters.

\subsection{Does Growing Strategy Lead to Distinct Behaviour?}
\begin{hypoanswer}[]
\textbf{Hypothesis.} Compared to \MD, \LD produces more symmetric weights and engages the attention sublayers in the central blocks more strongly, which we hypothesize contributes to its better empirical performance.

\medskip
\textbf{Evidence.} In \LD, inter-block cosine similarities are higher and more symmetric about the centre. Skipping the first attention sublayer in the middle blocks causes larger relative changes in the hidden state of the token under consideration.
\end{hypoanswer}

\textbf{Experiments.} To measure the weight similarity of blocks for the grown model, we concatenate all weights from the feedforward layers of a block and calculate the cosine similarity to other blocks. Similarly to before, we skip layers and measure the relative change for all later layers, but now on \emph{all} tokens (including the current token).

\begin{figure}[htb]
\centering
    \begin{subfigure}[b]{0.39\linewidth}
        \includegraphics[width=\linewidth]{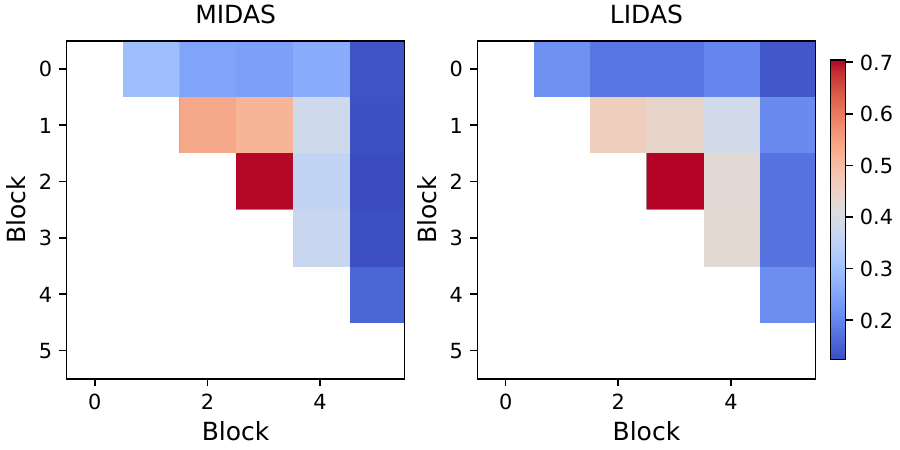}
        \caption{\centering Block similarity}
        \label{fig:main_block_similarity}
    \end{subfigure}%
    \begin{subfigure}[b]{0.61\linewidth}
    \centering
   \includegraphics[width=\linewidth]{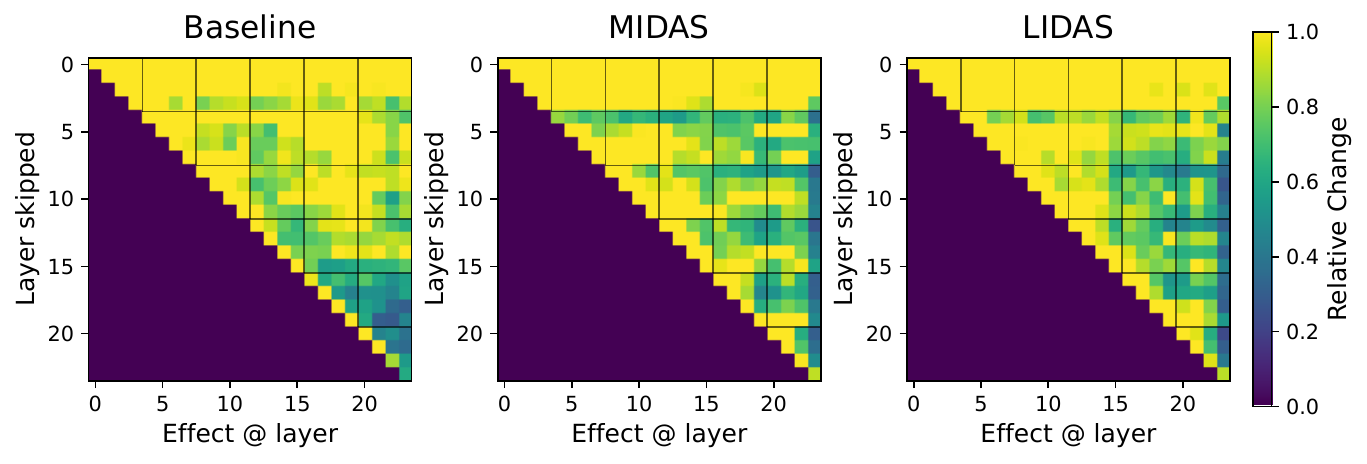}

    \caption{\centering Effect of skipping attention sublayer}
    \label{fig:current_attention_big}
    \end{subfigure}
\caption{\textbf{Baseline vs. \MD vs. \LD.} 
(a) The weight similarity, measured by cosine similarity, between feedforward layers per block is more symmetric for \LD compared to \MD. We omit the baseline as its weight similarities are all close to zero. 
(b) Skipping the first attention sublayer of every block in the centre of the network has a lower effect on the following layers' \emph{current} computations in \MD compared to \LD.
}
\label{fig:ana:mvl}
\end{figure}

\textbf{Interpretation.} In \LD we observe a block-similarity structure that is symmetric about the model’s centre, whereas in \MD the central block is more similar to the earlier (upper) blocks than to the later (lower) ones, yielding an asymmetric pattern (\cref{fig:ana:mvl}). This difference follows from the growth rule: \LD duplicates the exact layer-wise middle, while \MD is constrained to the nearest block centre. With an even number of blocks, the \MD choice necessarily biases similarity toward one side.

Additionally, this growing strategy leads to a higher utilisation of the first attention sublayer of every block (\cref{fig:ana:mvl}b), making it more aligned with the residual stream and having a bigger effect on the current computations of future layers. \changed{This effect is especially noticeable for deeper networks (Appendix \cref{fig:apx:small_current_attention}), but we also observe it here for the 1.7B model with 24 layers.}

\section{\changed{Discussion}}

\changed{\textbf{Mechanistic View.}
The results suggest that gradual-depth growing qualitatively changes how Transformers use their depth. In contrast to standard training, where late layers can often be removed with a modest performance decrease due to the ``Curse of Depth'' \citep{sun2025curse, csordas2025language}, depth-grown models allocate indispensable computation to later layers (\cref{sec:depth_utilization}). The layer-wise analyses indicate that layers in the middle of the network are not homogeneous, but organized into \emph{permutable, block-wise computations} (\cref{sec:comp_blocks}) with an internal \emph{cyclical (sub)layer structure} (\cref{sec:growth_patterns}). Taken together, the reported observations support a single picture: depth growth steers models toward learning a compact set of computational structures that are repeated along depth. This mirrors the spirit of Looped or Universal Transformers, which explicitly recursively apply a learned block of layers. However, here this loop-like behaviour emerges from the growth process without explicit weight tying. In this sense, the experiments support the hypothesis that depth grown models can be viewed as a relaxed version of Looped Transformers, where gradual growth (with layer duplication) steers the optimisation towards repeated computation without enforcing exact parameter sharing. }

\changed{\textbf{\MD\ vs.\ \LD.}
While \MD is motivated in \citet{saunshi2024inductive} by approximating the functional symmetries of Looped Transformers, its growing strategy is constrained by the block size and yields an asymmetric weight similarity pattern. We propose \LD that modifies the growing strategy to duplicate at the layer-wise middle, restoring more symmetric weights similarities (\cref{fig:main_block_similarity}) and therefore being closer to the original motivation of \MD. Quantitatively, we observe that \LD often further amplifies the strengths of \MD by improving in reasoning-heavy tasks (Math Word, Primitives), while negating its weaknesses by matching or exceeding the baseline in language modelling (NLL, Lambada, Hellaswag) at a 23\% reduced training cost (\cref{tab:exp:results}). Qualitatively, \LD aligns its attention sublayers better to the residual stream compared to \MD (\cref{fig:main_cyclical_attention}) making better use of the attention sublayers in the middle of the network (\cref{fig:current_attention_big}).}

\section{Conclusion}
This work systematically investigates how gradual depth growth in large language models affects their computational dynamics, providing a mechanistic explanation for their improved reasoning performance. Our findings confirm that gradually grown models outperform conventionally trained baselines on reasoning tasks and in training cost. Through detailed analysis, we demonstrate that this performance is tied to a more effective utilization of model depth. Unlike non-grown models that suffer from a Curse of Depth \citep{sun2025curse, csordas2025language}, our grown models continue to perform novel computations in their later layers and exhibit a higher overall depth score. We show that this is enabled by the formation of permutable computational blocks in the middle of the network, with each layer within these blocks serving a distinct cyclical role. The superiority of our proposed lightweight and novel stacking variant \LD is attributed to its ability to create a more symmetric weight structure and more effective attention layers, leading to improved empirical results. In conclusion, our research provides critical insights into the internal workings of depth-grown models, confirming that these training procedures can overcome key architectural inefficiencies and pave the way for more efficient and capable model development.

\section*{Acknowledgments and Disclosure of Funding}
The authors would like to thank João Sacramento for insightful discussions and support throughout this work.

This work was partially supported by the Helmholtz Foundation Model Initiative and the
Helmholtz Association. The authors gratefully acknowledge the Gauss Centre for Supercomputing e.V. (www.gauss-centre.eu) for funding this project by providing computing time through the John von Neumann Institute for Computing (NIC) on the GCS Supercomputer JUPITER | JUWELS \citep{JUWELS} at Jülich Supercomputing Centre (JSC). Furthermore, the authors appreciate the computational resources provided by the National High Performance Computing Centre (www.nhr.kit.edu).

 \clearpage

\if{0}
\subsubsection*{Author Contributions}
If you'd like to, you may include  a section for author contributions as is done
in many journals. This is optional and at the discretion of the authors.

\subsubsection*{Acknowledgments}
Use unnumbered third level headings for the acknowledgments. All
acknowledgments, including those to funding agencies, go at the end of the paper.
\fi

\bibliography{reference}
\bibliographystyle{abbrvnat}

\clearpage
\appendix

\section{SmolLM: Architecture \& Data}
\label{sec:apx:smollm}

\paragraph{Data.}
For all SmolLM models we trained, we followed the SmolLM-v1 data mixture from \cite{benallal2024smollmcorpus}.

\begin{itemize}
    \item \textbf{FineWeb-Edu (deduplicated)} \citep{benallal2024smollmcorpus}: Educational slice of FineWeb selected with a Llama3-70B–trained “educational quality” classifier. We use the deduplicated subset ($\approx$220B tokens) included in the SmolLM-Corpus.
    \item \textbf{OpenWebMath} \citep{paster2023openwebmath}: High-quality mathematical web pages extracted from Common Crawl with math-aware parsing, quality filtering, and deduplication ($\approx$14.7B tokens). Used to enrich math/reasoning coverage.
    \item \textbf{Cosmopedia v2} \citep{benallal2024smollmcorpus}: Synthetic textbooks, stories, and code generated with Mixtral-8×7B using curated topic lists and seed pages. v2 totals $\approx$39M documents ($\approx$28B tokens of textbooks/stories).
    \item \textbf{Python-Edu} \citep{benallal2024smollmcorpus}: Educational Python subset built by training an “educational code” classifier on annotated samples from \textit{The Stack} and applying it to the StarCoder training corpus. It contains $\approx$4B tokens with strict quality thresholding.
\end{itemize}
Given a fixed training-token budget, we then sample the corpus by proportion---70\% FineWeb-Edu (deduplicated), 15\% Cosmopedia v2, 9\% Python-Edu and 6\% OpenWebMath. Note that this leads to significant upsampling of the smaller datasets like Python-Edu and OpenWebMath.

\paragraph{Model architecture.}
Both sizes follow a LLaMA-style, decoder-only Transformer with \mbox{RMSNorm}, SwiGLU MLPs, and RoPE positional embeddings (tied input/output embeddings). The 360M variant uses GQA. 

\begin{table}[ht]
\centering
\begin{tabular}{lcc}
\toprule
 & \textbf{SmolLM-1.7B} & \textbf{SmolLM-360M} \\
\midrule
Layers & 24 & 32 \\
Model width & 2048 & 960 \\
FFN dimension & 8192 & 2560 \\
Attention heads & 32 & 15 \\
KV heads & 32 (MHA) & 5 (GQA) \\
Norm & RMSNorm & RMSNorm \\
MLP activation & SwiGLU & SwiGLU \\
Batch size & 2M & 1M \\
Learning rate $\eta_{\max}$ & 0.0005 & 0.003 \\
Weight decay & 0.01 & 0.01 \\
Positional embeddings & RoPE ($\theta{=}10{,}000$) & RoPE ($\theta{=}10{,}000$) \\
Context length (pretrain) & 2048 & 2048 \\
Tokenizer & cosmo2 & cosmo2 \\
Tied embeddings & Yes & Yes \\
\bottomrule
\end{tabular}
\caption{\textbf{Hyperparameters for both SmolLM models}}
\label{tab:apx:hparams}
\end{table}

\paragraph{Training.}
We train both sizes for 200k iterations. This corresponds to roughly 200B seen tokens for the 360M model and 400B for the 1.7B model. We use a trapezoidal learning-rate schedule with a linear warmup for the first 2000 steps up to the peak rate $\eta_{\max}$, a constant plateau until step 170000, and a 1-\text{sqrt} decay over the final 30000 steps \citep{hagele2024scaling}. We optimise with AdamW \citep{loshchilov2019decoupledweightdecayregularization} and apply global gradient clipping at 1.0 for all runs.

\paragraph{Training with the Growing operator}
For SmolLM with gradual depth growth all training hyperparameters match the baseline in \cref{tab:apx:hparams}. We use a fixed block size $b=4$ and insert a new \emph{middle} block after each stage, instantiating either MIDAS (duplicate the middle stage block) or LIDAS (duplicate the layer-wise middle; see \cref{sec:growing_operator}), while keeping width and attention heads constant. At every growth step we deep-copy all layer parameters and their optimizer state so duplicated layers start identically (same AdamW moments) and then diverge with continued training; embeddings and the final head are copied unchanged. The number of growth stages is defined by $k = L_{\text{final}}/b$ and $T$ is the total number of training steps. We allocate per-stage budgets using the PROP-$\alpha$ schedule of \citet{saunshi2024inductive}:
\[
  T_i \;=\; \frac{i^{\alpha}}{\sum_{j=1}^{k} j^{\alpha}}\, T \quad \text{for } i=1,\dots,k\,,
\]
and use PROP-1 ($\alpha{=}1$) in our experiments. In practice, we round $T_i$ to integers (largest-remainder to keep $\sum_i T_i=T$) and maintain a \emph{single continuous} learning-rate schedule across stages (no LR reset; the scheduler’s global step carries over). We set $T=170{,}000$ so all models reach their final depth before they enter the cooldown phase.
 
\paragraph{Compute requirements.} We trained all models on NVIDIA A100 GPUs (40 GB). The large (static baseline) model ran on 128 GPUs for 4.5 days, and the small (static baseline) model ran on 64 GPUs for 1.5 days.

\section{Evaluation Setup}
\label{sec:apx:exp_setup}
\paragraph{Codebases.}
Depth analyses and interventions follow the methodology of \citet{csordas2025language}, which is extended to block-wise skip/swap operations over consecutive layers (block sizes $\{1,2,4,8\}$) and further extended to permuting consecutive layers in arbitrary order. Tuned Lens experiments follow \citet{belrose2023eliciting}.

\paragraph{Reproducibility defaults.}
We adopt the default configuration from the original depth-analysis repository of \citet{csordas2025language} for reproducibility. Specifically, we use the same fixed set of GSM8K prompts/examples for early-exit, skip, swap and relative contribution evaluations, and we keep random seeds, batching, and evaluation hyperparameters at their defaults unless stated otherwise.

\paragraph{Models and data.}
We analyze SmolLM-v1 backbones at 360M and 1.7B parameters (training details in \cref{sec:apx:smollm}) and evaluate on MATH, MQuAKE, and GSM8K as described in the main text. Preprocessing follows \citet{csordas2025language}.

\paragraph{Intervention protocols.}
\label{par:apx:intervention_protocol}
We distinguish \emph{heatmap (relative-change) experiments} from \emph{benchmarked interventions}. Heatmaps quantify relative changes and use \textbf{single (sub)layer skipping only}. Benchmarked interventions (accuracy-based) are described separately below.
For heatmaps, we evaluate two intervention \emph{modes} and two \emph{measurement axes}, following and extending \citet{csordas2025language}:
\begin{itemize}
  \item \textbf{Current vs. future effects.} In the \emph{current} setting, we intervene by erasing the entire (sub)layer contribution \emph{for all tokens} and measure changes on all positions. In the \emph{future} setting, for a chosen boundary token index $t$, we erase the (sub)layer contribution \emph{only for tokens $\le t$}, leaving tokens $>t$ unchanged at that (sub)layer; we then measure changes \emph{strictly on tokens $>t$}. This design directly tests whether information is transferred to later tokens via attention, ruling out purely pointwise (self-only) computation.
  \item \textbf{Output vs. later-layer effects.} For the \emph{output probability distribution}, we compute the L2 norm difference between the softmaxed logits of the intervened and original forward passes, aggregated over the relevant positions (current or future). For the \emph{later-layer effects}, we compute, for each later layer, the \emph{relative change} in the residual contribution (i.e., the norm of the difference in that layer’s residual update divided by the norm of the original residual update), again aggregated over the relevant positions.
\end{itemize}
Concretely for heatmaps, in the \emph{future} effects evaluation we select multiple boundary indices $t$ and, for each $t$, (i) erase the (sub)layer’s contribution only at tokens $\le t$, (ii) keep its contribution intact at tokens $>t$, and then compare the intervened and original runs on (a) softmaxed output distributions at positions $>t$ and (b) residual contributions of all later layers at positions $>t$. This directly tests whether features are moved forward in time (to future tokens) by attention.

For heatmaps, we also include a \textbf{local (direct) effects} variant, which isolates pairwise dependencies between a source layer and a later target layer without allowing effects to \emph{propagate} through multiple subsequent layers. Specifically, for a source layer $s$ and a later layer $\ell>s$, we subtract the stored contribution of $s$ from the residual fed into $\ell$ and record the relative change at $\ell$; we do \emph{not} roll this modification forward beyond $\ell$. This complements the propagated analyses by revealing direct, non-compounded influences.

Heatmap interventions are performed at the layer or sublayer level and are \textbf{strictly single-layer}. The current/future distinction applies \emph{only} to these heatmap experiments. Block-wise operations are used solely in benchmarked interventions (below).

\paragraph{Aggregation for heatmaps.}
For heatmap visualizations of later-layer effects, we aggregate by taking the \emph{maximum} relative change across (i) batch examples, (ii) eligible sequence positions, and (iii) multiple chosen boundaries $t$ in the future setting. Concretely, for current effects, we take the max over all positions; for future effects, we take the max only over positions strictly greater than $t$, and then take the max over all tested $t$ for each example. This yields a single matrix of source-layer by target-layer maxima per model/setting.

\paragraph{Tuned Lens training and evaluation.}
Following \citet{belrose2023eliciting}, we train, for each layer, a small affine adapter that maps that layer’s residual output to the hidden representation with the same shape that serves as the input to the \emph{final} normalization layer immediately before the unembedding. Final logits are then obtained by applying the model’s final normalization and unembedding as usual. Adapters are trained on a held-out split of FineWeb-Edu and evaluated by (a) KL divergence between early-exit and final distributions and (b) top-5 vocabulary overlap with the final prediction (cf. \cref{fig:midas:depth_efficiency}B/C for 1.7B and \cref{fig:midas:depth_efficiency_360}B/C 360M).

\paragraph{Benchmarked interventions.}
We evaluate accuracy on downstream benchmarks under: (i) Tuned Lens early-exit (using the adapter path described above), (ii) skip interventions, and (iii) swap interventions. For benchmarks, we may intervene on contiguous \textbf{blocks} of sizes $\{2,4,8\}$ (in addition to single layers). We decode with greedy top-1 and compute benchmark accuracy (e.g., Math Word, reasoning primitives), matching the evaluation protocol used for the unmodified model. The current/future distinction does \emph{not} apply to benchmark evaluations.

\paragraph{Depth score.}
We report the \emph{logit-effect} depth score based on \texttt{mean\_dout}. For each layer $\ell$, \texttt{mean\_dout} is the across-examples mean of the maximum L2 change in the softmaxed output distribution at future tokens when intervening at layer $\ell$ (future-setting; see intervention protocols). We normalize this per-layer vector to a probability distribution over layers and take its expected layer index as the depth score \citep{csordas2025language}. 

\section{Detailed Benchmark Results}
\label{sec:apx:detailed_results}

\changed{\paragraph{Setup} In \cref{sec:exp_benchmarks} we presented aggregated results over several Benchmarks. In this section, we show the detailed results for all models and for completeness, we also report results with \LN  and further experiment with the combination of \LN and \LD / \MD. We evaluated all models on these benchmarks using the language model evaluation harness library \citep{eval-harness}.}

\paragraph{Reasoning Primitives} We implemented the \textbf{Reasoning Primitives} following the task descriptions in \citet{saunshi2024inductive}. Induction copying is generated by sampling a sequence of random 3-letter words (e.g., length 10), selecting a contiguous subsequence (e.g., length 5) from within it, appending that subsequence, and asking for the next token in the original sequence. Variable assignment is generated by sampling variable--value statements and querying a single variable’s value. We use the same basic, math and code prompt templates.

\changed{An example for a \textit{Copying random words} task would be:}

\textit{Prompt}: \begin{verbatim}
    Fill in the blank:
    jic dqy sof uzg ewr oxw osp tkj rvw mnu jic dqy sof uzg ewr ___. ->
\end{verbatim}

\textit{Answer}: \begin{verbatim}
    oxw
\end{verbatim}

For an example of \textit{variable assignment} task:

\textit{Prompt}: \begin{verbatim}
Fill in blank:

o=23
k=3
t=13
a=1
e=9
o=___. ->
\end{verbatim}

\textit{Answer:}
\begin{verbatim}
    23
\end{verbatim}

Notice that for the above tasks multiple choice format is used and a 5-shot evaluation setting. This means that the random guessing baseline score is $10\%$ for the Copying task and $20\%$ for the variable assignment task. 

\paragraph{Results} 
\begin{figure}[ht]
    \begin{minipage}[b]{1.0\textwidth}
     \resizebox{0.9999\textwidth}{!}{ 
        \addtolength{\tabcolsep}{-0.2em}
        \begin{tabular}{ll|ccccc|c}
        \toprule
        \multicolumn{2}{c|}{\bf } & \multicolumn{1}{r}{\bf CoQA} & \multicolumn{1}{r}{\bf DROP} & \multicolumn{1}{r}{\bf QuAC} & \multicolumn{1}{r}{\bf SquadV2} & \multicolumn{1}{r}{\bf TyDi QA (wc)} & \multicolumn{1}{r}{\bf Mean} \\
        \toprule
        360M & Baseline & 46.08 & 12.48 & 14.27 & 24.35 & 17.25 & 22.89 \\
        \cmidrule(l){2-8}
        & \MD & 50.00 & 12.75 & 14.10 & 24.96 & 21.06 & 24.57 \\
        & \LD & 51.50 & \textbf{15.25} & \textbf{15.79} & 28.11 & \textbf{22.50} & \textbf{26.63} \\
        & \changed{\LN} & 44.80 & 13.18 & 12.45 & 24.11 & 21.14 & 23.14 \\
        & \changed{\LN + \MD} & 45.70 & 13.14 & 13.86 & 25.11 & 12.39 & 22.04 \\
        & \changed{\LN + \LD} & \textbf{53.17} & 13.03 & 14.59 & \textbf{30.69} & 16.19 & 25.5 \\
        \toprule
        1.7B & Baseline & 58.36 & 16.52 & 15.91 & 33.88 & 23.17 & 29.57 \\
        \cmidrule(l){2-8}
        & \MD & 59.35 & 16.88 & 17.30 & 36.06 & 14.39  & 28.80\\
        & \LD & \textbf{63.41} & 17.66 & \textbf{17.91} & \textbf{36.56} & 13.65 & \textbf{29.84}\\
        & \changed{\LN} & 54.94 & \textbf{17.84} & 16.61 & 32.78 & \textbf{23.37} & 29.11\\
        & \changed{\LN + \LD} & 62.36 & 16.09 & 17.74 & 34.98 & 9.05 & 28.04\\
        \bottomrule
        \end{tabular}
        }
        \small \captionof{table}
        {\textbf{Open-book QA Benchmarks}}.
        \label{tab:apx:open_book}
    \end{minipage}
\end{figure}

In \cref{tab:apx:open_book,tab:apx:closed_book} we report per-dataset results for Open-Book and Closed-Book QA. In line with \citet{saunshi2024inductive}, both grown models (\MD and \LD) yield larger gains on Open-Book QA than on Closed-Book QA. Notably, \LD\ 1.7B improves over the 1.7B baseline even on most Closed-Book datasets and remains competitive on the rest, which differs from observations made for \MD in \citet{saunshi2024inductive}. \changed{Overall, the grown variants confer modest but consistent Open-book gains, whereas \LN alone yields only small changes relative to the non-grown baseline. Combining growing with \LN can sometimes improve over the standard \LN setting, but often still falls short when compared to \LD.} 

\begin{figure}[ht]
    \begin{minipage}[b]{1.0\textwidth}
     \resizebox{0.9999\textwidth}{!}{ 
        \addtolength{\tabcolsep}{-0.2em}
            \begin{tabular}{ll|cccc|c}
            \toprule
            \multicolumn{2}{c|}{\bf } & \multicolumn{1}{r}{\bf Trivia QA} & \multicolumn{1}{r}{\bf Web Questions} & \multicolumn{1}{r}{\bf TyDi QA (nc)} & \multicolumn{1}{r}{\bf Natural Questions} & \multicolumn{1}{r}{\bf Mean} \\
            \toprule
            360M & Baseline & 19.23 & \textbf{16.78} & 12.98 & 9.01 & 14.50 \\
            \cmidrule(l){2-7}
            & \MD & 18.90 & 14.58 & 12.64 & 8.89 & 13.75 \\
            & \LD & \textbf{20.80} & 15.61 & 12.14 & 9.73 & 14.57 \\
            & \changed{\LN} & 20.52 & 16.73 & 12.36 & \textbf{9.94} & \textbf{14.89} \\
            & \changed{\LN + \MD} & 17.76 & 16.60 & 12.66 & 9.18 & 14.05 \\
            & \changed{\LN + \LD} & 18.23 & 15.32 & \textbf{13.67} & 9.26 & 14.12 \\
            \toprule
            1.7B & Baseline & 27.72 & 19.20 & 15.34 & 12.18 & 18.61 \\
            \cmidrule(l){2-7}
            & \MD & \textbf{27.98} & 17.96 & 16.16 & 11.91 & 18.50\\
            & \LD & 26.85 & 20.24 & \textbf{16.34} & \textbf{12.90} & \textbf{19.08} \\
            & \changed{\LN} & 25.12 & \textbf{20.89} & 15.71 & 12.79 & 18.63\\
            & \changed{\LN + \LD} & 25.88 & 20.15 & 15.78 & 11.88 & 18.42\\
            \bottomrule
            \end{tabular}
            }
            \small \captionof{table}{\textbf{Closed-book QA Benchmarks}.}
            \label{tab:apx:closed_book}
    \end{minipage}
\end{figure}

\begin{figure}[ht]
    \begin{minipage}[b]{1.0\textwidth}
     \resizebox{0.9999\textwidth}{!}{ 
        \addtolength{\tabcolsep}{-0.2em}
        \begin{tabular}{ll|cccccc|c}
        \toprule
        \multicolumn{2}{c|}{\bf } & \multicolumn{1}{r}{\shortstack[c]{\textbf{ASDiv}}} & \multicolumn{1}{r}{\shortstack[c]{\textbf{MAWPS} \\ Add/Sub}} & \multicolumn{1}{r}{\shortstack[c]{\textbf{MAWPS} \\ Multi-Arith}} & \multicolumn{1}{r}{\shortstack[c]{\textbf{MAWPS} \\ Single-Op}} & \multicolumn{1}{r}{\shortstack[c]{\textbf{MAWPS} \\ Single-Eq}} & \multicolumn{1}{r}{\bf SVAMP} & \multicolumn{1}{r}{\bf Mean}\\
        \toprule
        360M & Baseline & 3.34 & 3.67 & \textbf{1.72} & 5.66 & 2.75 & 5.02 & 3.69 \\
        \cmidrule(l){2-9}
        & \MD & 3.77 & 3.67 & 1.15 & 6.29 & 6.42 & 5.02 & \textbf{4.39} \\
        & \LD & \textbf{4.64} & 1.83 & \textbf{1.72} & \textbf{7.55} & 6.42 & 4.01 & 4.36 \\
        & \changed{\LN} & 2.95 & 0.00 & 1.15 & 4.40 & 5.50 & 3.34 & 2.89 \\
        & \changed{\LN + \MD} & 3.56 & \textbf{5.50} & 1.15 & 1.89 & 6.42 & \textbf{5.69} & 4.04 \\
        & \changed{\LN + \LD} & 3.21 & 3.67 & \textbf{1.72} & 4.40 & \textbf{7.34} & 3.68 & 4.00 \\
        \toprule
        1.7B & Baseline & 11.15 & 14.68 & 1.15 & 25.16 & \textbf{22.02} & 8.36 & 13.75 \\
        \cmidrule(l){2-9}
        & \MD & 12.93 & 18.35 & 2.30 & 33.96 & 20.18 & 8.70 & 16.07\\
        & \LD & \textbf{14.88} & \textbf{25.69} & 2.87 & \textbf{38.36} & 18.35 & 11.37 & \textbf{18.59}\\
        & \changed{\LN} & 10.20 & 13.76 & 1.72 & 17.61 & 14.68 & 8.03 & 11.00\\
        & \changed{\LN + \LD} & 14.19 & 22.02 & \textbf{3.45} & 31.45 & 21.10 & \textbf{11.71} & 17.32\\
        \bottomrule
        \end{tabular}
        }
        \small \captionof{table}{\textbf{Math Word}.}
        \label{tab:apx:math_world}
    \end{minipage}
\end{figure}

On the reasoning benchmarks, Math Word (\cref{tab:apx:math_world}) and Reasoning Primitives (\cref{tab:apx:primitives}), improvements at 360M are modest on average, while at 1.7B they become more pronounced. For Math Word, \LD\ 1.7B attains the best score on five out of six benchmarks (the exception is MAWPS Single-Equation). \changed{For Reasoning Primitives, both \MD and \LD\ surpass the baseline, with \LD\ 1.7B leading on copying and on the code/math variable-assignment formats, while \MD\ slightly edges \LD\ on the basic variable-assignment format. We notice however, that the variance of performance between tasks is much higher compared to language tasks.}

\begin{figure}[ht]
    \begin{minipage}[b]{1.0\textwidth}
     \resizebox{0.9999\textwidth}{!}{ 
        \addtolength{\tabcolsep}{-0.2em}
        \begin{tabular}{ll|ccccc|c}
        \toprule
        \multicolumn{2}{c|}{\bf } & \multicolumn{1}{r}{\shortstack[c]{\textbf{Copying}\\ \\(random words)}} & \multicolumn{1}{r}{\shortstack[c]{\textbf{Copying}\\ \\(real words)}} & \multicolumn{1}{r}{\shortstack[c]{\textbf{Variable} \\ \textbf{assignment} \\ (basic)}} & \multicolumn{1}{r}{\shortstack[c]{\textbf{Variable} \\ \textbf{assignment} \\ (code)}} & \multicolumn{1}{r}{\shortstack[c]{\textbf{Variable} \\ \textbf{assignment} \\ (math)}} & \multicolumn{1}{r}{\shortstack[c]{\textbf{Mean}\\ \\ \phantom{assignment} }}\\
        \toprule
        360M & Baseline & 15.50 & 13.30 & 20.50 & 58.30 & 42.70 & 30.06 \\
        \cmidrule(l){2-8}
        & \MD & 13.80 & 14.10 & 20.00 & 52.90 & 40.10 & 28.18\\
        & \LD & 14.20 & 19.70 & 24.30 & 51.80 & 46.00 & 31.20\\
        & \changed{\LN} & 17.90 & 16.40 & 22.70 & 49.10 & 50.80 & 31.38 \\
        & \changed{\LN + \MD} & 15.80 & 16.00 & \textbf{26.30} & \textbf{58.70} & 48.40 & 33.04 \\
        & \changed{\LN + \LD} & \textbf{21.90} & \textbf{19.80} & 25.50 & 57.20 & \textbf{52.10} & \textbf{35.30}\\
        \toprule
        1.7B & Baseline & 16.80 & 23.60 & 20.80 & 64.20 & 48.80 & 34.84 \\
        \cmidrule(l){2-8}
        & \MD & 19.30 & 24.60 & \textbf{37.00} & 61.50 & 62.00 & 40.88\\
        & \LD & \textbf{28.40} & \textbf{31.00} & 36.70 & 71.80 & 68.80 & \textbf{47.34} \\
        & \changed{\LN} & 24.40 & 26.60 & 23.80 & \textbf{75.20} & \textbf{71.90} & 44.38\\
        & \changed{\LN + \LD} & 17.00 & 23.00 & 34.80 & 71.20 & 70.60 & 43.32\\
        \bottomrule
        \end{tabular}
        }
        \small \captionof{table}{\textbf{Reasoning Primitives}.}
        \label{tab:apx:primitives}
    \end{minipage}
\end{figure}
\changed{Consistent with our depth analyses, these benchmark trends coincide with higher depth scores and later-layer reliance for grown models, whereas LN-Scaling in our setup does not increase depth utilization relative to the baseline nor improve performance.}

In addition to improved reasoning performance, models trained with gradual stacking also require fewer computational resources, as reported in previous works. Specifically, \MD and \LD only need $\approx 77\%$ of the FLOPs used to train the baseline in \cref{tab:apx:flops}.
\begin{table}[]
    \centering
        \begin{tabular}{lrr}
        \toprule
        Model & PetaFLOPs & Ratio \\
        \midrule
        360M Standard & 613527.488 & 1.289 \\
        360M Grown & 476147.897 & 1.000 \\
        1700M Standard & 4813222.102 & 1.288 \\
        1700M Grown & 3736608.182 & 1.000 \\
        \bottomrule
        \end{tabular}
    \caption{\textbf{PetaFLOPs used for training 200k iterations with block size of 4 and PROP-1 growing schedule.}}
    \label{tab:apx:flops}
\end{table}

\section{\changed{Additional results on depth Analysis}}
In \cref{sec:depth_analysis} we have shown how growing can alter the structure within the model, leading to better depth utilisation and different altering of the residual stream and robustness. However, these results have mainly been presented for \MD on the 1.7B model scale. In this section we want to first add results for \LD, to show that it does exhibit the same patterns as \MD and also show results on the smaller 360M scale. Finally, we include additional ablation plots for block sizes different from 4.

\subsection{1.7B Models}
\begin{figure*}[t]
  \centering
  \includegraphics[width=\textwidth]{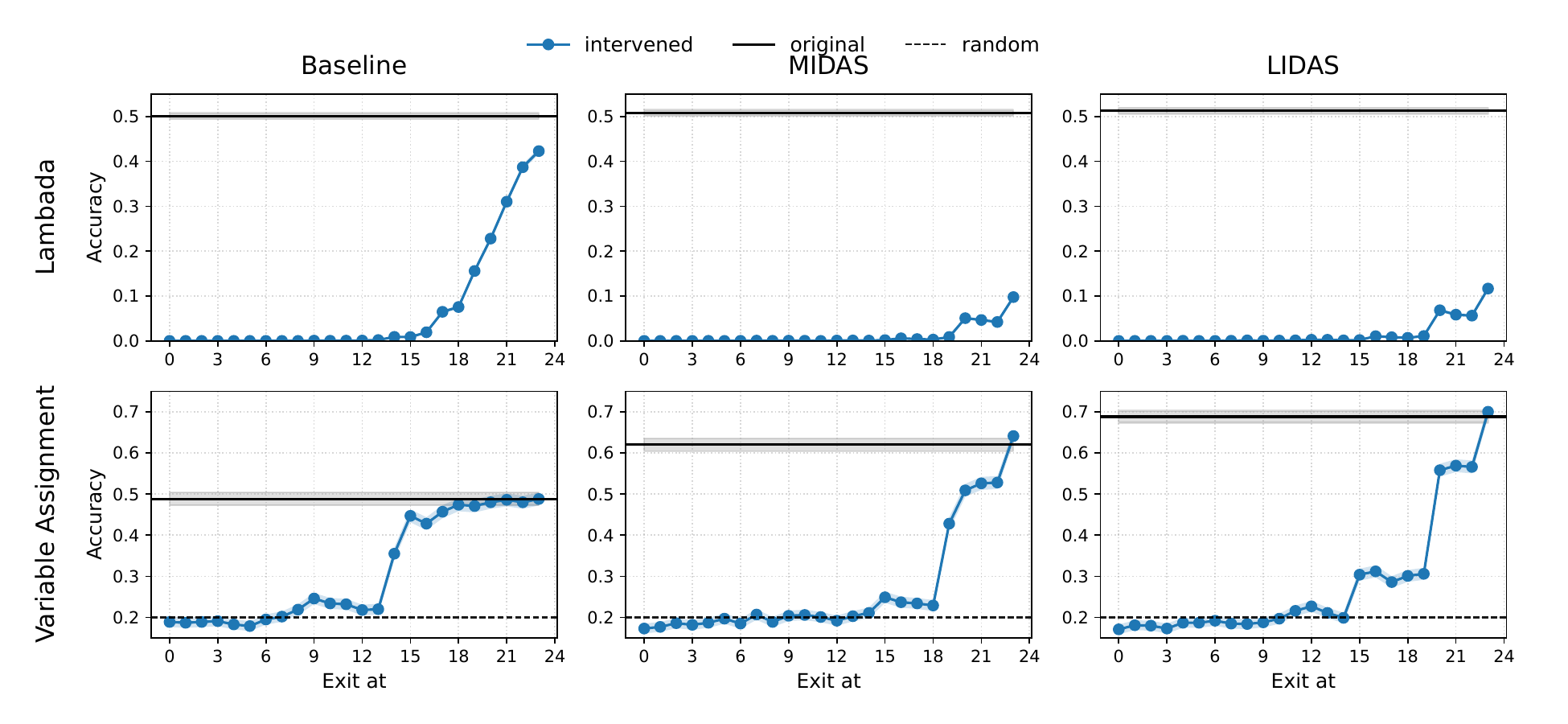}
  \caption{Early exit with tuned lens on \textit{Lambada} and \textit{Variable Assignment Math} for the Baseline, \MD, and \LD models at scale 1.7B.}
  \label{fig:apx:big_early_exit_tuned_lens}
\end{figure*}

This section extends the main analyses for the 1.7B models to \LD. For each hypothesis made in \cref{sec:depth_analysis}, we show that our results also hold for \LD by resuming each experiment.
\label{sec:supp_big_models}
\paragraph{Additional results for \cref{sec:depth_utilization}}
\cref{fig:apx:big_early_exit_tuned_lens} reproduces the early-exit analysis at 1.7B on Lambada and Variable Assignment, showing that grown models' early exit performance is relatively poor over the entire stack while the baseline saturates much earlier. Notably, this result is stable across tasks with different absolute accuracies, suggesting that reliance on later layers reflects a training-induced computational pattern rather than task difficulty. This complements the \cref{sec:depth_utilization} diagnostics and reinforces that depth growth yields genuinely deeper computation at scale.

\begin{figure*}[t]
  \centering
  \includegraphics[width=\textwidth]{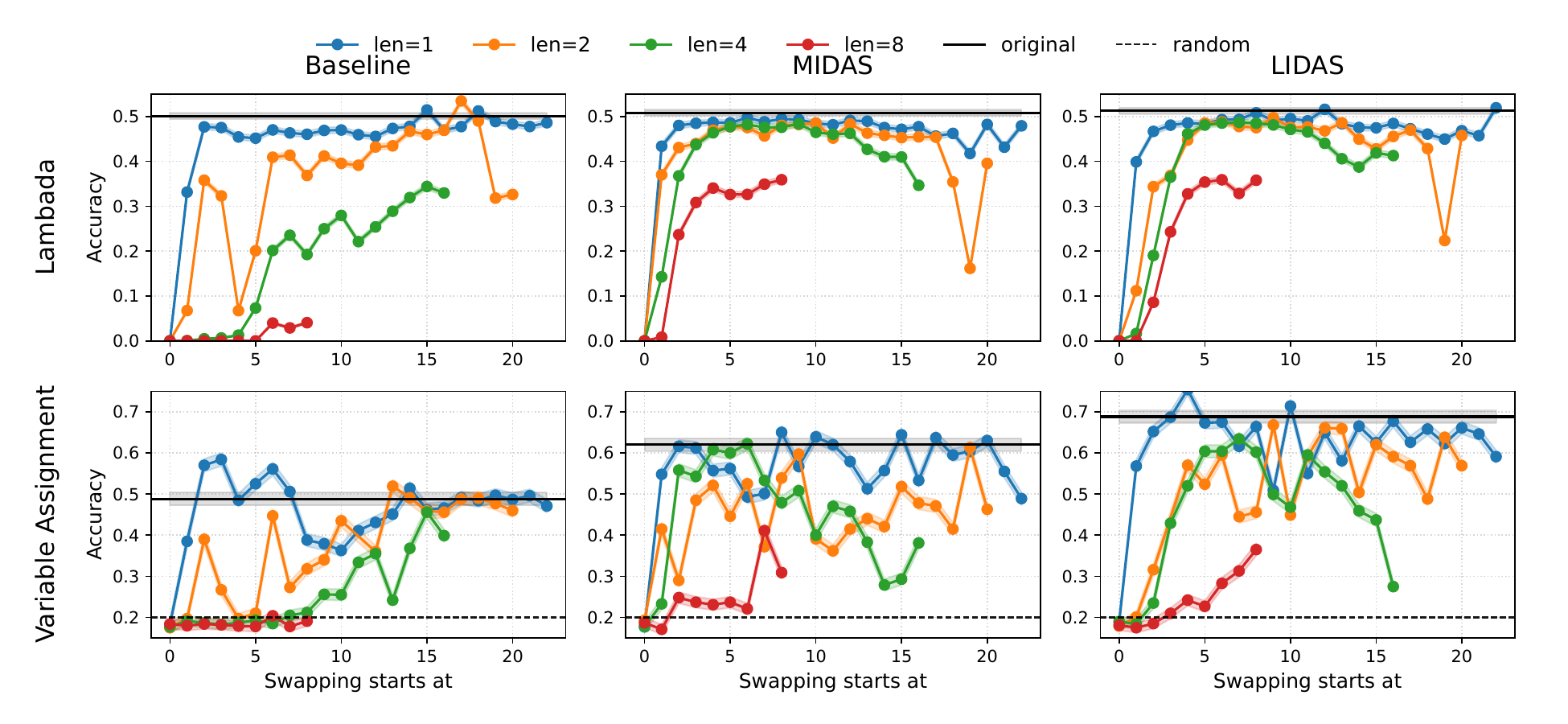}
  \caption{Swap ablations on \textit{Lambada} and \textit{Variable Assignment Math} for the Baseline, \MD, and \LD models at scale 1.7B.}
  \label{fig:apx:big_swap}
\end{figure*}

\begin{figure*}[t]
  \centering
  \includegraphics[width=\textwidth]{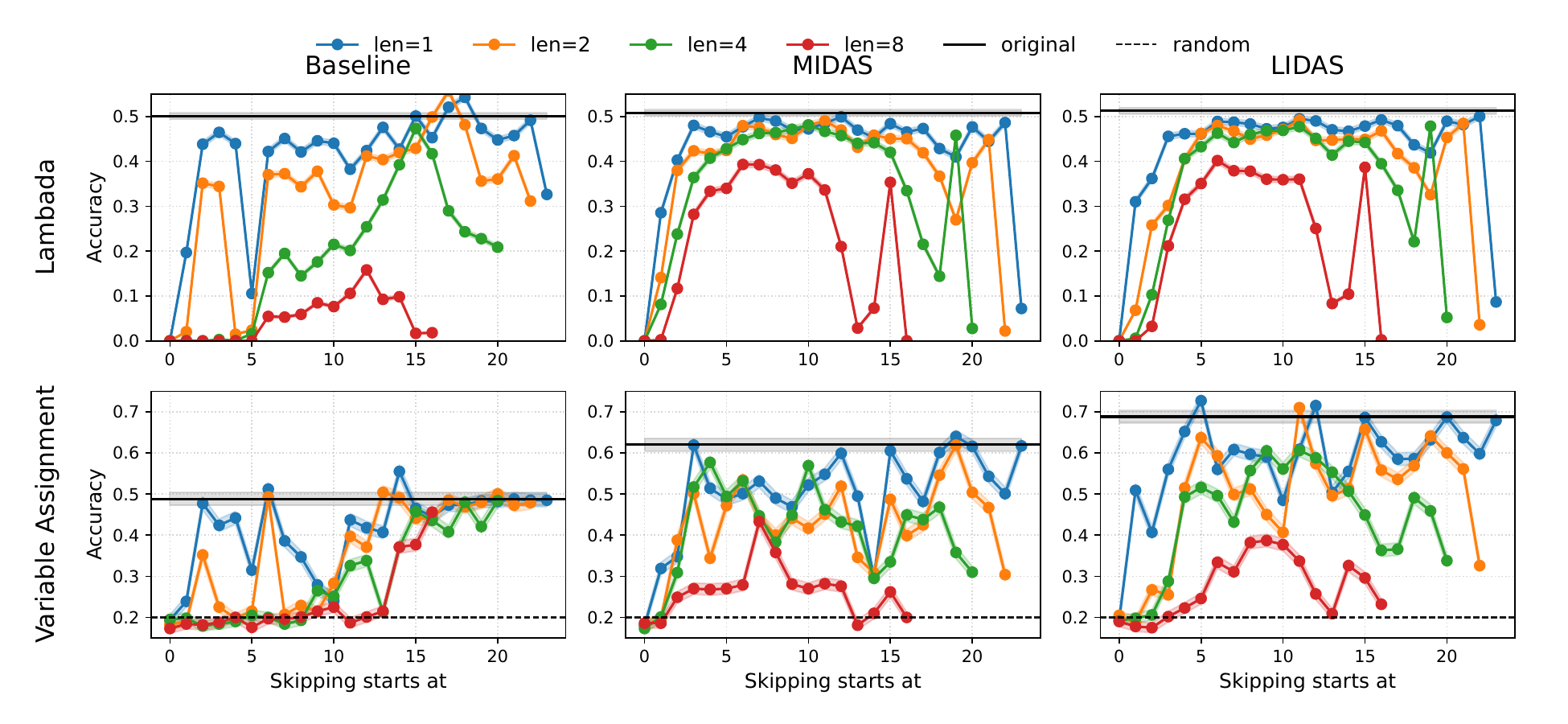}
  \caption{Skip ablations on \textit{Lambada} and \textit{Variable Assignment Math} for the Baseline, \MD, and \LD models at scale 1.7B.}
  \label{fig:apx:big_skip}
\end{figure*}

\paragraph{Additional results for \cref{sec:comp_blocks}}
\cref{fig:apx:big_swap} extends the swap-ablation result (\cref{fig:main_skipping_swapping_lambada}) to include \LD at 1.7B. Grown models are markedly more robust than the baseline when swapping multi-layer blocks (sizes 2–8), especially in the middle of the network, conforming to the signature of block-level permutability argued in \cref{sec:comp_blocks}. \cref{fig:apx:big_skip} provides the complementary experiment, in which we skip consecutive layers of different sizes. Together, these interventions support our hypothesis that depth growth organizes computation into mid-network blocks whose presence is crucial but whose internal order is comparatively flexible.

\begin{figure*}[t]
  \centering
  \includegraphics[width=\textwidth]{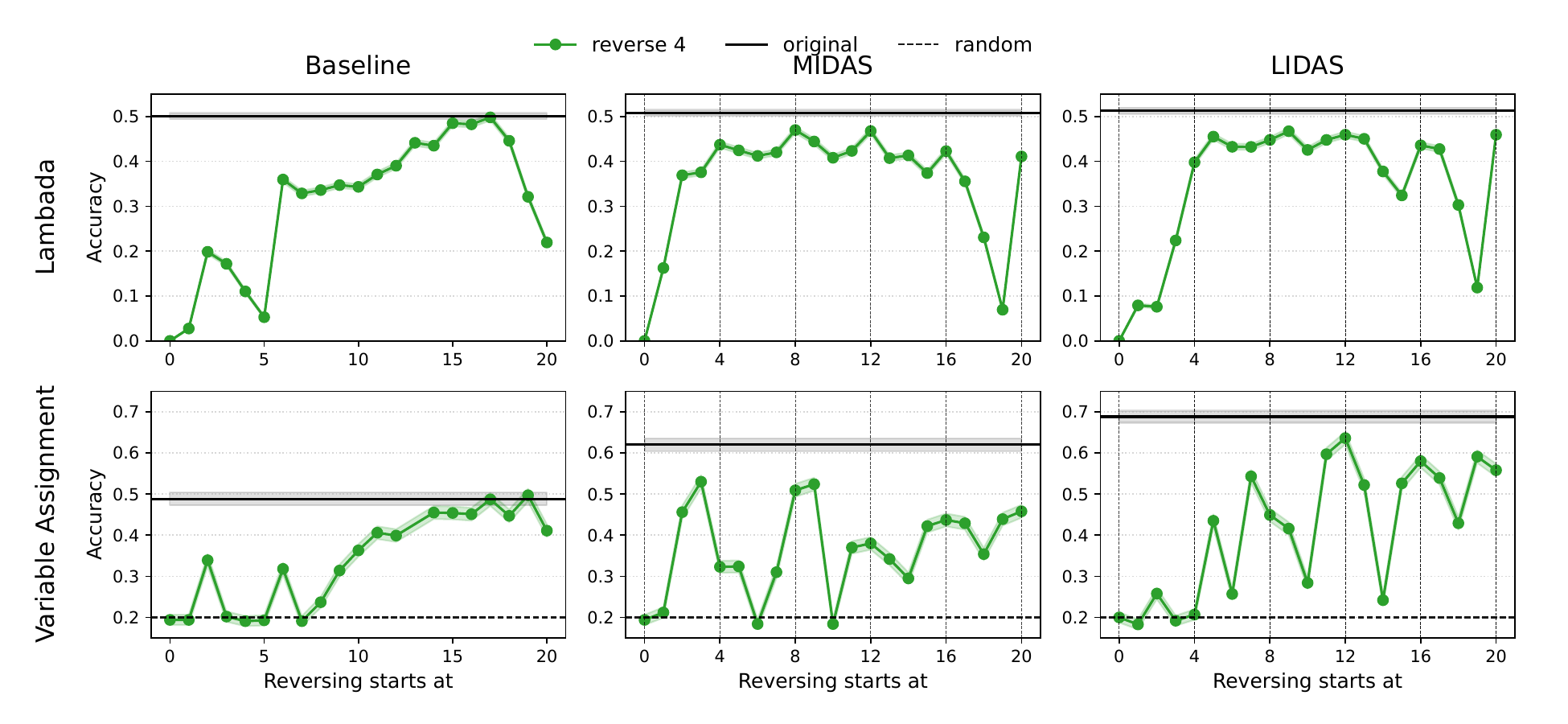}
  \caption{Reversing the order of 4 consecutive layers on \textit{Lambada} and \textit{Variable Assignment Math} for the Baseline, \MD, and \LD models at scale 1.7B.}
  \label{fig:apx:big_reverse}
\end{figure*}

\begin{figure*}[t]
  \centering
  \includegraphics[width=\textwidth]{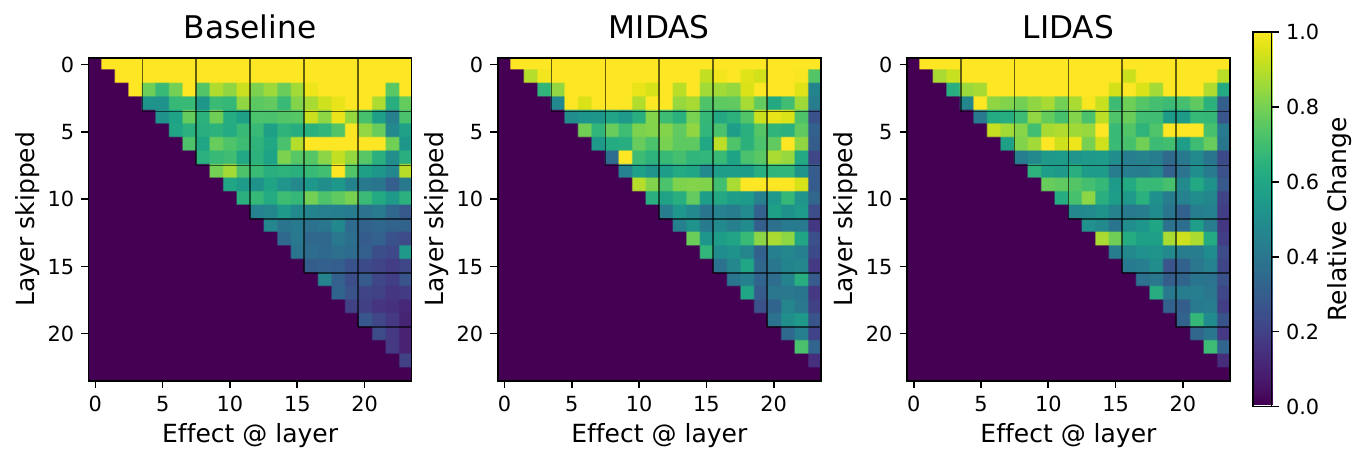}
  \caption{Propagated future effects of single-layer skipping for the Baseline, \MD, and \LD models at scale 1.7B.}
  \label{fig:apx:big_future_effects}
\end{figure*}

\begin{figure*}[t]
  \centering
  \includegraphics[width=\textwidth]{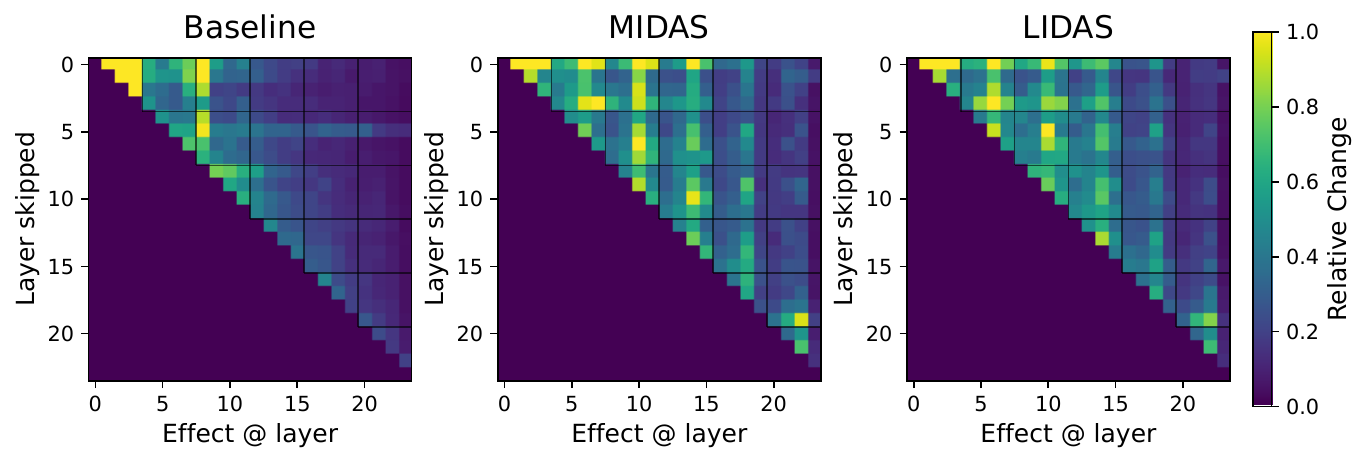}
  \caption{Local future effects of single-layer skipping for the Baseline, \MD, and \LD models at scale 1.7B.}
  \label{fig:apx:big_future_local_effects}
\end{figure*}

\paragraph{Additional results for \cref{sec:growth_patterns}}
Figure \cref{fig:apx:big_reverse} shows that grown models are particularly fragile when reversing four-layer windows that pass block boundaries, degrading more than under swapping or skipping, indicating that while blocks are permutable as units, the intra-block progression encodes roles that do not commute, as argued in \cref{sec:growth_patterns}. We can clearly see that \LD also follows this pattern and that it is independent of the task being considered. \cref{fig:apx:big_future_effects,fig:apx:big_future_local_effects} reveal repeating mid-block motifs and stronger downstream propagation from later layers in grown models, generalising the results from \cref{fig:main_midas_heatmaps} to include \LD at 1.7B. Collectively, these results support that our claims from \cref{sec:growth_patterns} do hold for \LD and are not task-specific.

\subsection{360M Models}
\label{sec:supp_small_models}

This section presents the depth analysis results for the 360M models for the experiments described in the main paper. 

\paragraph{Benchmark results on different seed}
To test the robustness of our benchmark results to random initialisation and data ordering, we retrain the baseline 360M model and grown variants with a different model initialisation and data seed, while keeping all other hyperparameters fixed. As shown in \cref{tab:app:results_add_seed}, absolute scores change only slightly compared to \cref{tab:exp:results} and the relative ranking of methods is preserved: both MIDAS and LIDAS match the baseline on language-modelling and knowledge benchmarks, and LIDAS continues to achieve the strongest performance on Math Word and Reasoning Primitives. This indicates that our main conclusions about the benefits of gradual depth growth are stable across seeds.

\begin{figure}[thp!]
    \begin{minipage}[b]{1.0\textwidth}
     \resizebox{0.9999\textwidth}{!}{ 
        \addtolength{\tabcolsep}{-0.2em}
        \begin{tabular}{ll|c|cc|cc|cc|cc}
        \toprule
        \multicolumn{2}{c|}{} &\multicolumn{7}{c|}{Standard cooldown} &\multicolumn{2}{c}{Math cooldown}  \\ \cmidrule(){3-11}
        \multicolumn{2}{c|}{\bf } &\multicolumn{1}{c|}{\bf} &\multicolumn{1}{c}{\bf Open-book} &\multicolumn{1}{c|}{\bf Closed-book}&\multicolumn{1}{c}{\bf} &\multicolumn{1}{c|}{\bf} &\multicolumn{2}{c|}{} &\multicolumn{2}{c}{}  \\
        \multicolumn{2}{c|}{\bf } &\multicolumn{1}{c|}{\bf Holdout Set} &\multicolumn{1}{c}{\bf Q\&A} &\multicolumn{1}{c|}{\bf Q\&A}&\multicolumn{1}{c}{\bf Lambada} &\multicolumn{1}{c|}{\bf Hellaswag } &\multicolumn{1}{c}{\bf Math Word} &\multicolumn{1}{c|}{\bf Primitives } &\multicolumn{1}{c}{\bf Math Word} &\multicolumn{1}{c}{\bf Primitives } \\
        \multicolumn{2}{c|}{\bf } &\multicolumn{1}{c|}{(NLL $\downarrow$)} &\multicolumn{1}{c}{(F1 $\uparrow$)} &\multicolumn{1}{c|}{(F1 $\uparrow$)} &\multicolumn{1}{c}{(Acc $\uparrow$)} &\multicolumn{1}{c|}{(Acc $\uparrow$)} &\multicolumn{1}{c}{(Acc $\uparrow$)} &\multicolumn{1}{c|}{(Acc $\uparrow$)} &\multicolumn{1}{c}{(Acc $\uparrow$)} &\multicolumn{1}{c}{(Acc $\uparrow$)} \\
        \toprule
        \parbox[t]{2mm}{\multirow{4}{*}{\rotatebox[origin=c]{90}{360M}}} & Baseline &2.18 &23.18 &14.22 &43.16 & 40.16 &3.11 &29.92 & 7.91 & 37.36  \\
        \cmidrule(){2-11}
        &\MD          &2.18 &24.26 &\textbf{14.34} &42.58 &40.11 &\textbf{3.47} &34.08 &8.50 & 41.86\\
        & \LD   &\textbf{2.16} &\textbf{25.02} &14.08 &\textbf{44.27} &\textbf{40.90} &2.59 &\textbf{37.14} &\textbf{10.47} &\textbf{46.88} \\
        
        \bottomrule
        \end{tabular}
        }
        \small \captionof{table}{\textbf{Performance of the 360M baseline and depth-grown models under a different seed.} We keep the data mixture and training hyperparameters fixed and only change the seed of the random initialisation of model parameters and data order. The results closely match those in \cref{tab:exp:results}, confirming that the gains of MIDAS and especially LIDAS on reasoning benchmarks are robust to the choice of seeds.}
        \label{tab:app:results_add_seed}
    \end{minipage}
\end{figure}

\begin{figure}[ht]
    \vspace{-5mm}
    \begin{center}
        \includegraphics[width=\textwidth]{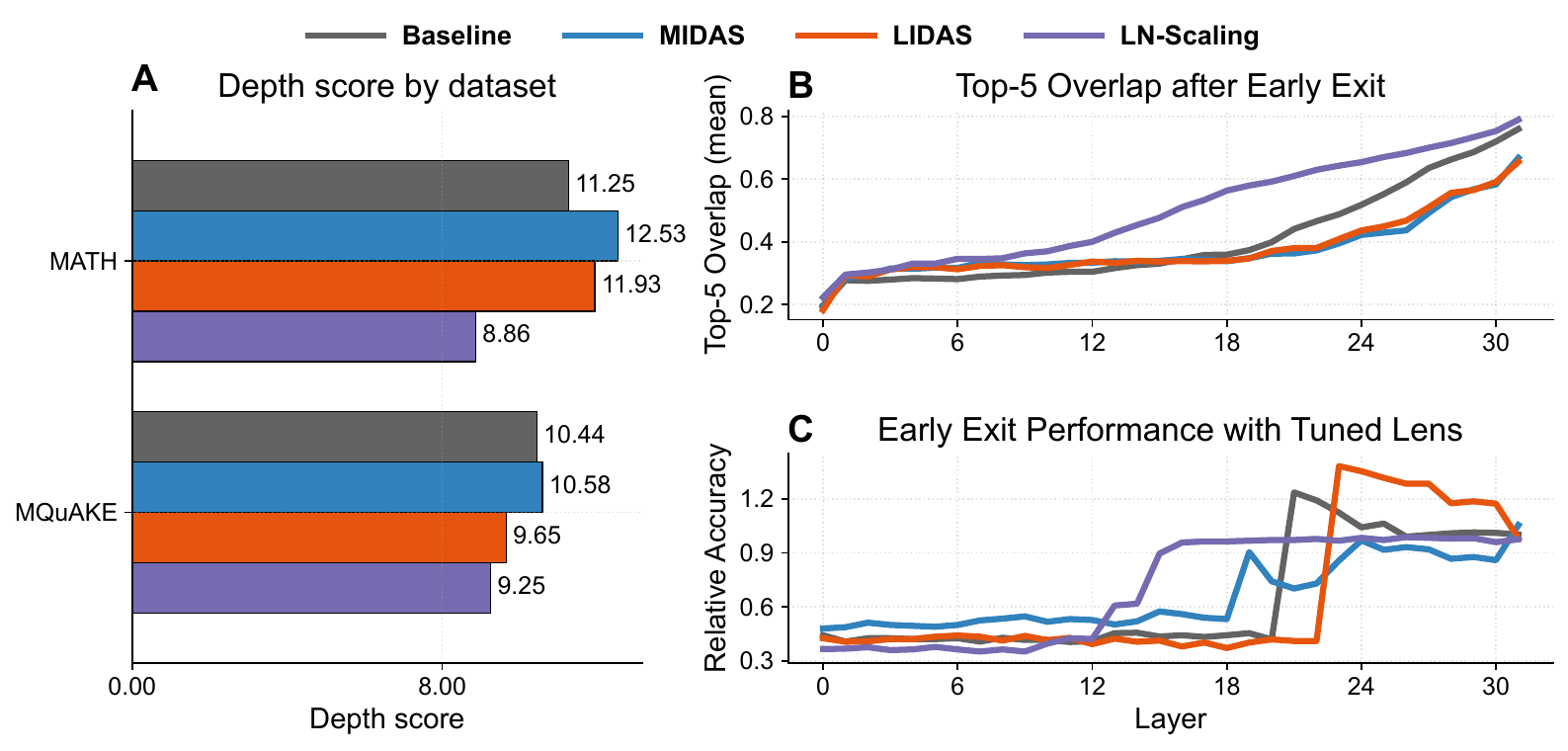}
    \end{center}
    \caption{ \textbf{Depth-grown models use their depth more (360M)}. (A) Depth score \citep{csordas2025language} on MATH \citep{hendrycks2021measuring} and MQuAKE \citep{zhong2023mquake}. Grown models (\MD, \LD) have consistently higher depth scores, except \LD on MQuAKE. (B) Top-5 overlap between each layer’s early-exit vocabulary and model’s final vocabulary on 20 prompts from GSM8K \citep{cobbe2021training}. Both grown models studied in this work exhibit lower overlap at later layers, indicating that these later layers still contribute additional features necessary for the final prediction. (C) Early-exit relative accuracy versus layer on \emph{Variable Assignment Math} reasoning primitive. The baseline reaches its best performance early, whereas accuracy for \MD and \LD is the highest at later layers. \changed{Using these metrics, however, \LN shows no discernible benefit over the baseline in depth utilisation.}}
    \label{fig:midas:depth_efficiency_360}
\end{figure}

\paragraph{Additional results for \cref{sec:depth_utilization}} In \cref{fig:midas:depth_efficiency_360}, 
we summarize 360M model depth utilization using the depth score and tuned-lens early-exit diagnostics (see \cref{fig:midas:depth_efficiency} for the 1.7B case). In \cref{fig:midas:depth_efficiency_360} A, we observe that the results are more task dependent compared to the 1.7B model: for experiments conducted on the MATH dataset, the trend of grown models to utilize more depth is evident (although to a lesser extent compared to the 1.7B model). However, for the MQuAKE dataset, the depth utilization pattern appears more complex, as only \MD has a higher depth score than the baseline, and no clear conclusion can be derived. In \cref{fig:midas:depth_efficiency_360} B, early exiting for the baseline model saturates much earlier, consistent with the 1.7B case. Moreover, in \cref{fig:midas:depth_efficiency_360} C we can also see that peak performance is reached earlier for baseline compared to \MD and \LD.

\label{sec:apx:360_results}

\begin{figure*}[t]
\centering
  \includegraphics[width=\textwidth]{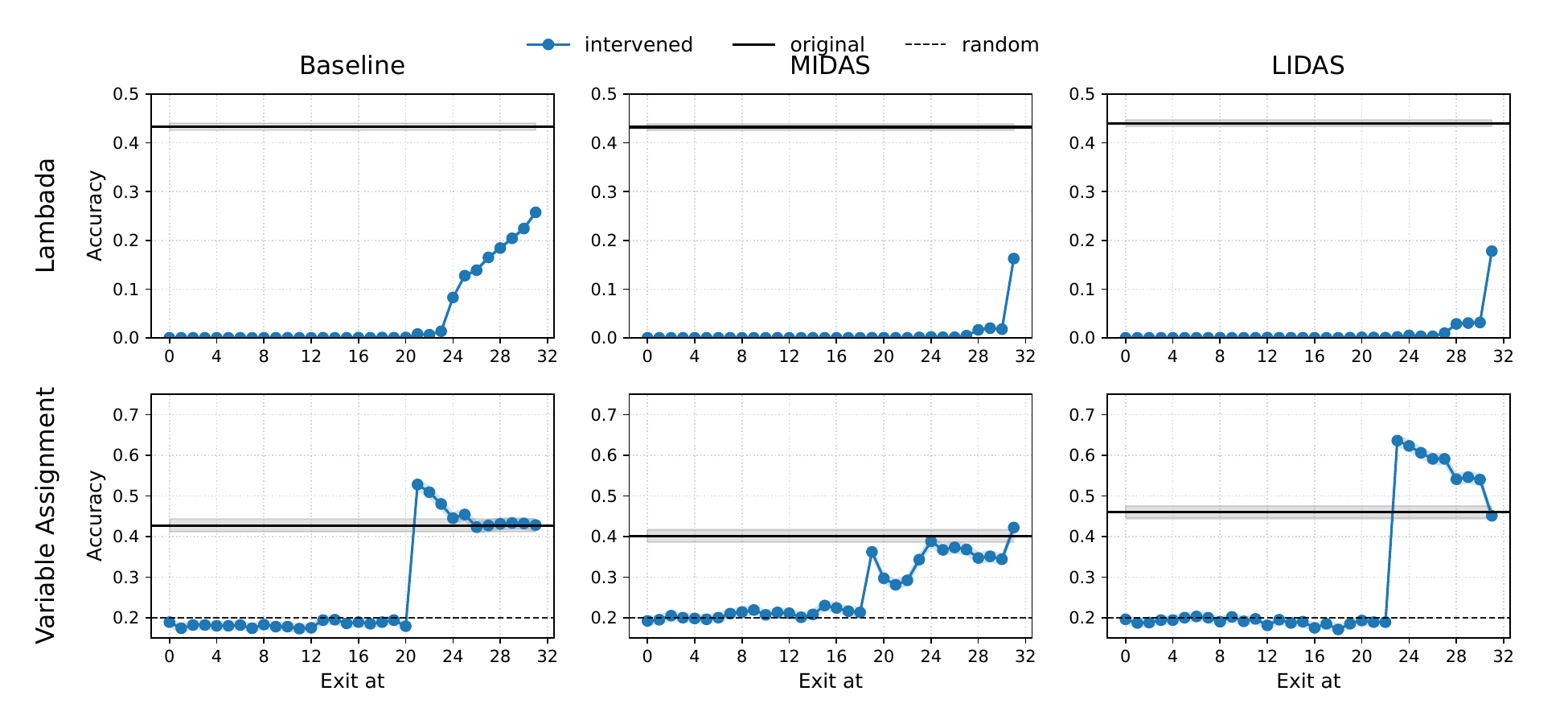}
  \caption{Small models (360M): early exit with tuned lens on \textit{Lambada} and \textit{Variable Assignment Math} for Baseline, \MD, and \LD.}
  \label{fig:apx:small_early_exit_tuned_lens}
\end{figure*}

In \cref{fig:apx:small_early_exit_tuned_lens} we replicate benchmarked early exit interventions at 360M: again the picture is less clear compared to the 1.7B model case and the results are task dependent; in Lambada early exiting performance for grown models stays close to 0 until the very end, indicating the necessity of later layers in information processing. In Variable Assignment, however, \MD and \LD exhibit different behaviour when compared to the baseline.

\begin{figure*}[t]
    \centering
  \includegraphics[width=\textwidth]{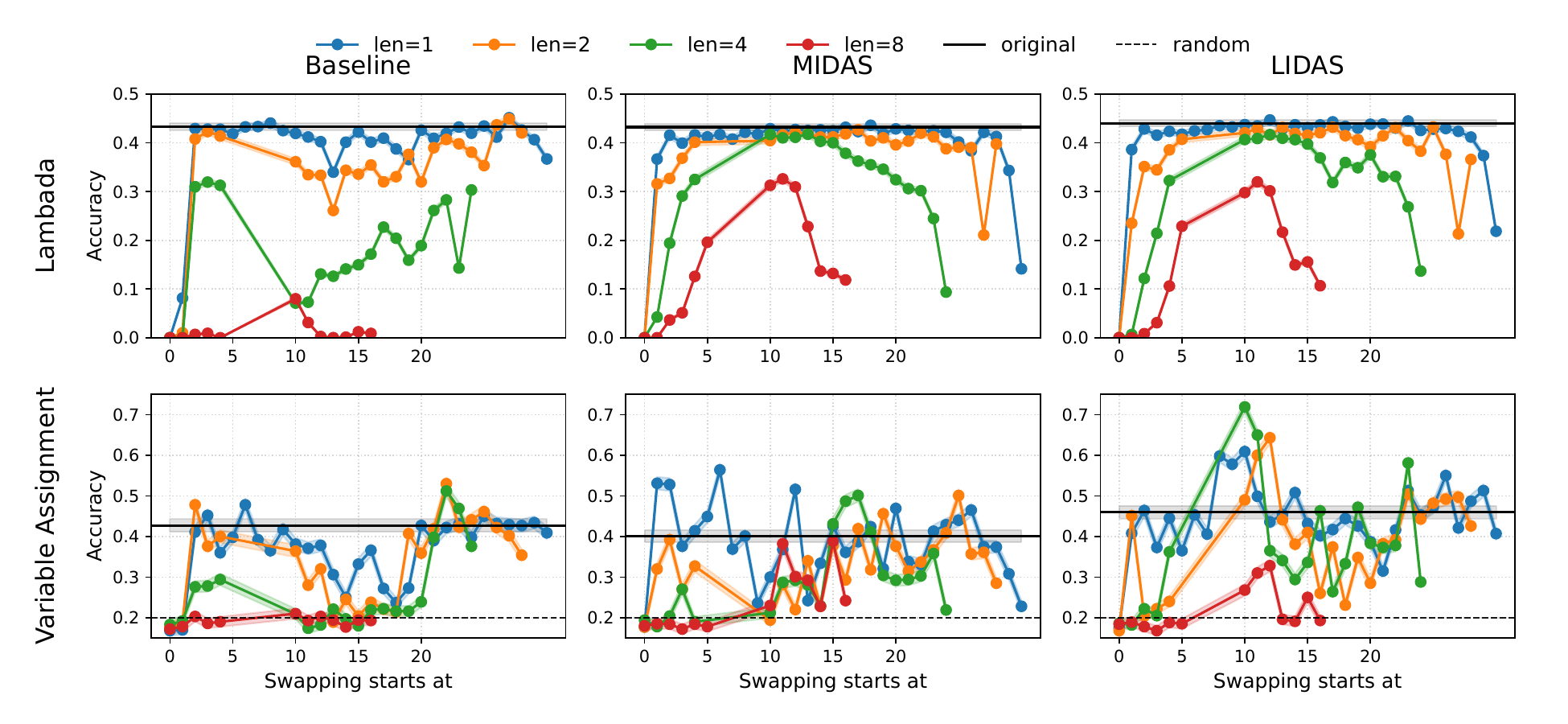}
  \caption{Small models (360M): swap ablations on \textit{Lambada} and \textit{Variable Assignment Math}.}
  \label{fig:apx:small_swap}
\end{figure*}

\begin{figure*}[t]
        \centering
  \includegraphics[width=\textwidth]{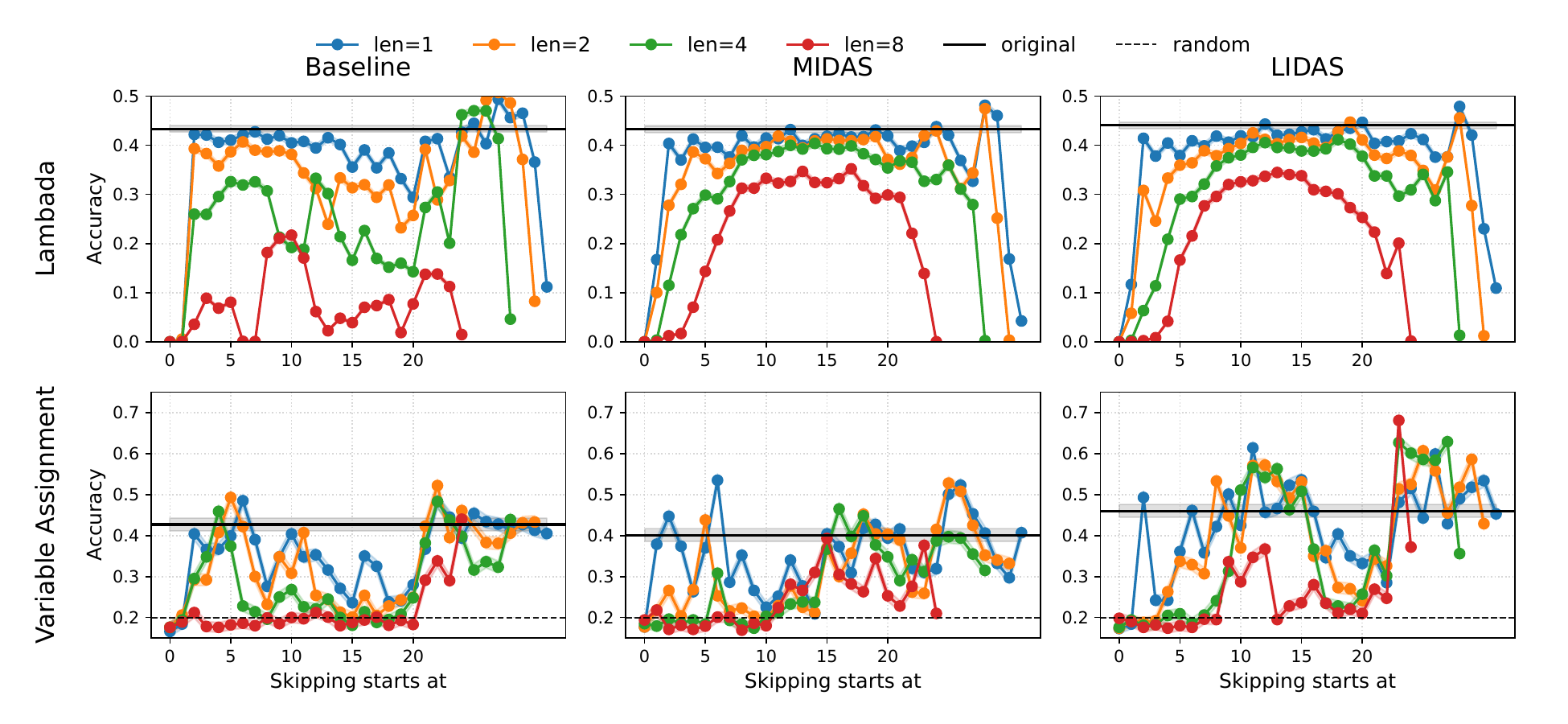}
  \caption{Small models (360M): skip ablations on \textit{Lambada} and \textit{Variable Assignment Math}.}
  \label{fig:apx:small_skip}
\end{figure*}

\begin{figure*}[t]
  \centering
  \includegraphics[width=\textwidth]{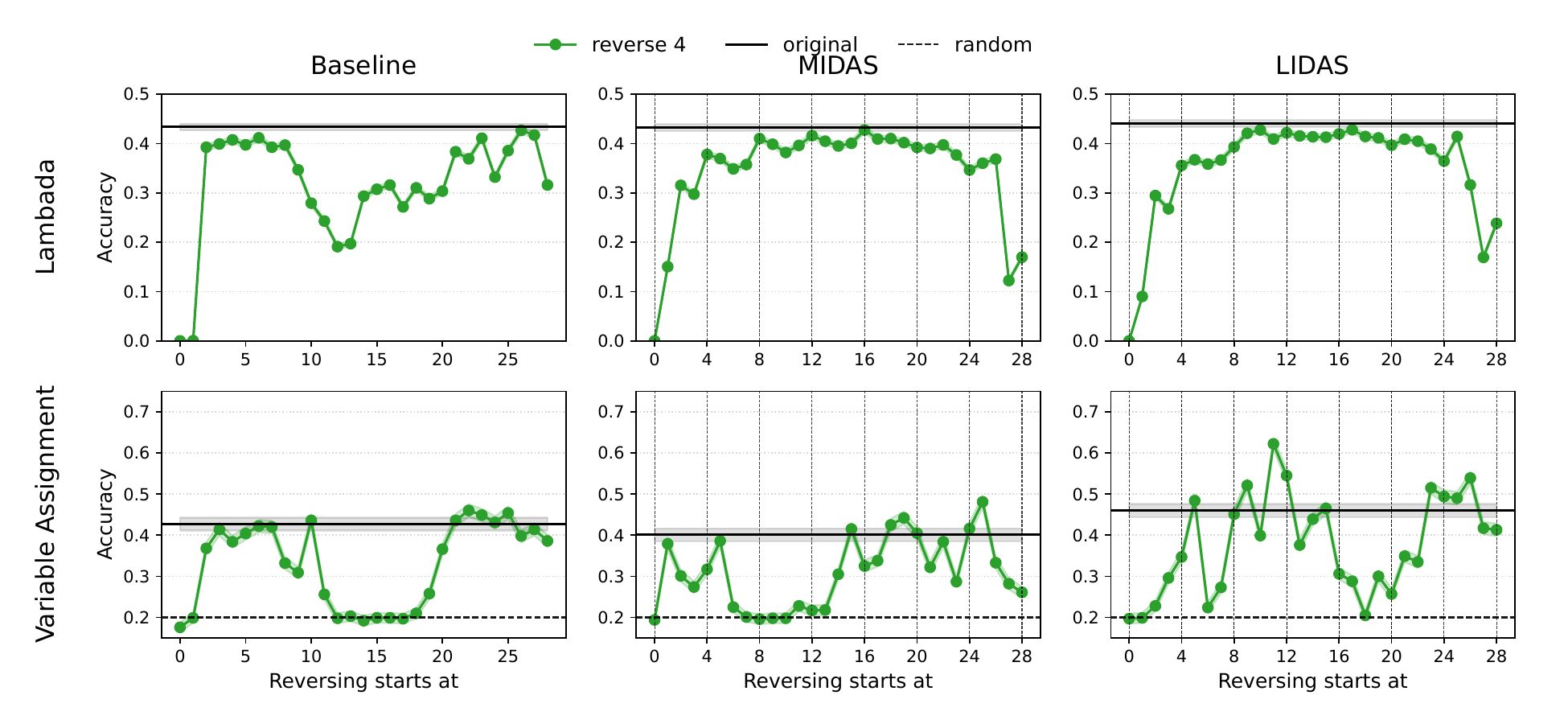}
  \caption{Small models (360M): reversing the order of 4 consecutive layers on \textit{Lambada} and \textit{Variable Assignment Math}.}
  \label{fig:apx:small_reverse}
\end{figure*}

\paragraph{Additional results for \cref{sec:comp_blocks}} In \cref{fig:apx:small_swap,fig:apx:small_skip,fig:apx:small_reverse} we assess robustness under reduced capacity, covering swap, skip, and reversal interventions. Consistent with the 1.7B case, the swap experiments show higher robustness w.r.t block level ordering interventions for the grown models. Moreover, when it comes to the reverse ordering interventions, we observe again this increased sensitivity of the grown models wrt block boundaries.

\begin{figure*}[t]
\centering
  \includegraphics[width=\textwidth]{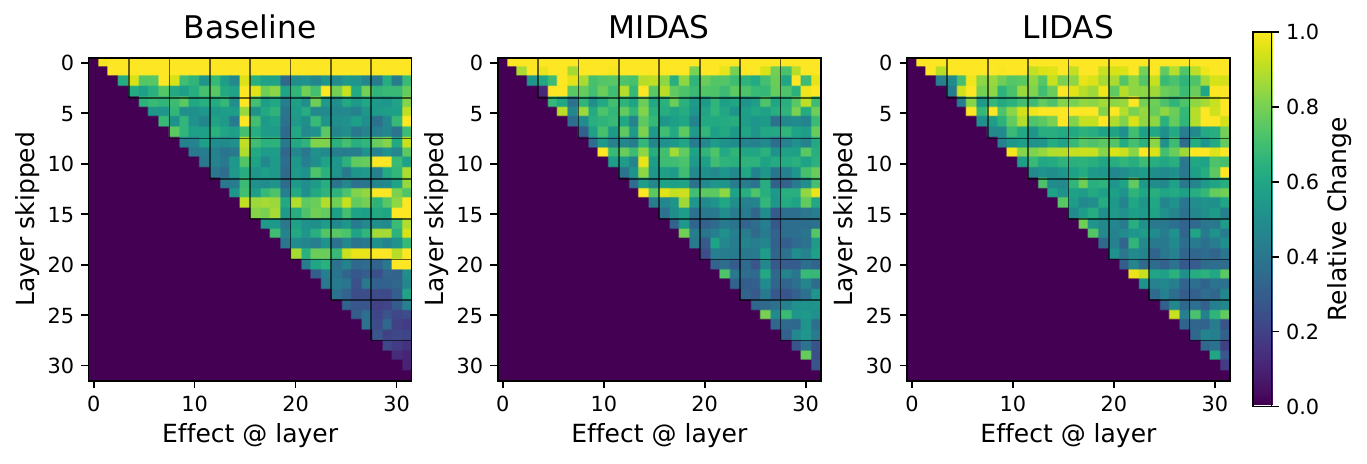}
  \caption{Small models (360M): propagated future effects of single-layer skipping.}
  \label{fig:apx:small_future_effects}
\end{figure*}

\begin{figure*}[t]
    \centering
  \includegraphics[width=\textwidth]{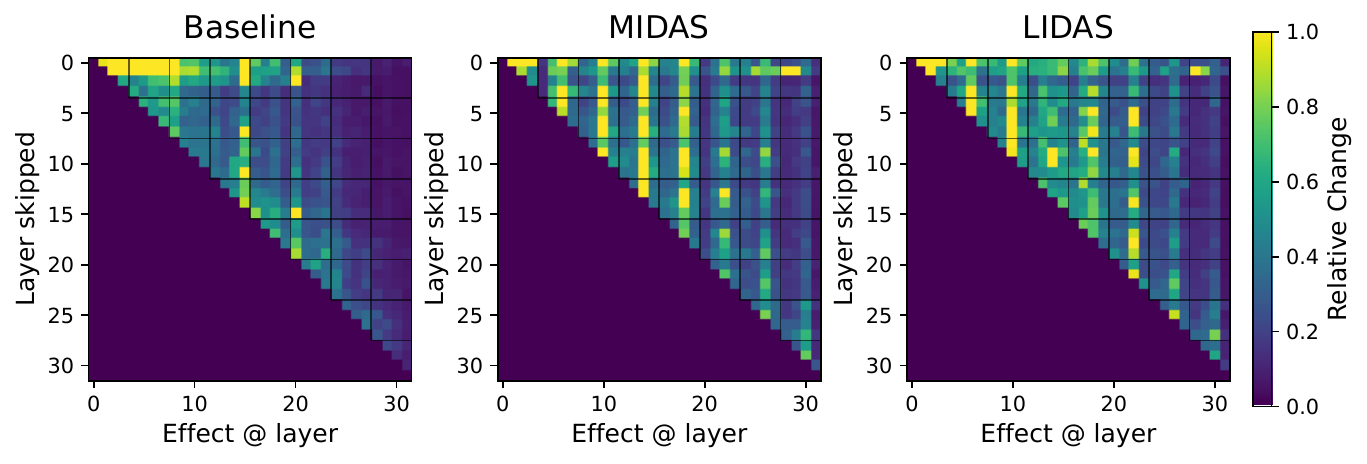}
  \caption{Small models (360M): local future effects of single-layer skipping.}
  \label{fig:apx:small_future_local_effects}
\end{figure*}

\begin{figure*}[t]
    \centering
  \includegraphics[width=\textwidth]{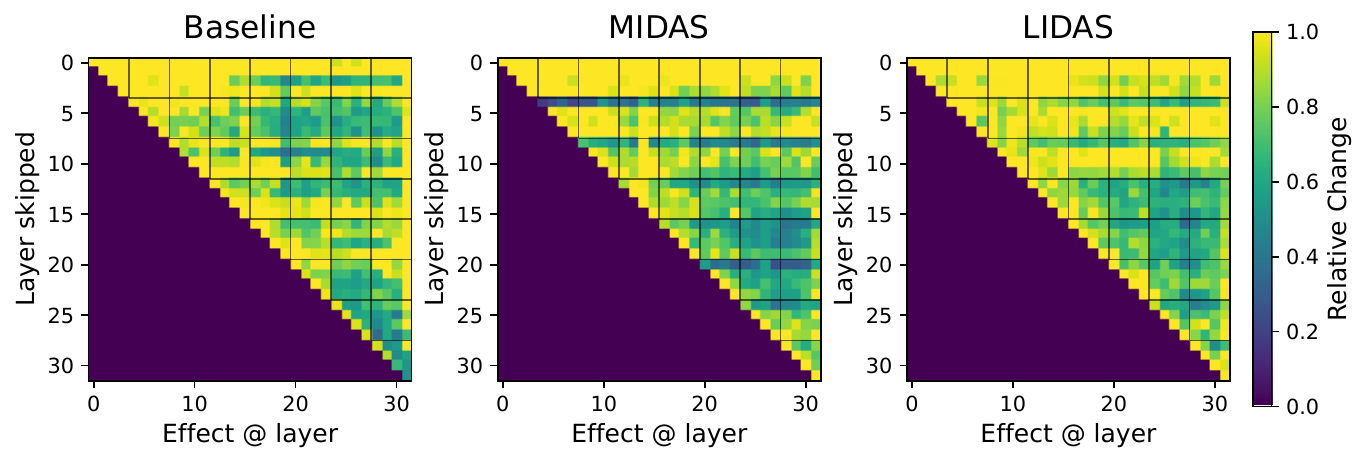}
  \caption{Small models (360M): current effects when skipping the attention sublayer.}
  \label{fig:apx:small_current_attention}
\end{figure*}

\paragraph{Additional results for \cref{sec:growth_patterns}} We include the small-model counterparts of the future propagated, future local, and current attention ablations (see \cref{fig:apx:small_future_effects,fig:apx:small_future_local_effects,fig:apx:small_current_attention}). In a nutshell, all cyclical patterns observed for the 1.7B model in the main paper still hold for the 360M case: the sensitivity of the 3rd layer of each block to the output of all its previous layers \cref{fig:apx:small_future_effects,fig:apx:small_future_local_effects} and also reduced impact of the first attention sublayer of every block to later layers for \MD model. Notice that these effects are even more pronounced for the 360M model compared to the 1.7B case (compare the corresponding light and dark stripes in \cref{fig:apx:small_future_local_effects,fig:apx:small_current_attention} to these in \cref{fig:apx:big_future_local_effects,fig:current_attention_big})

\begin{figure*}
\centering
\includegraphics[width=0.8\textwidth]{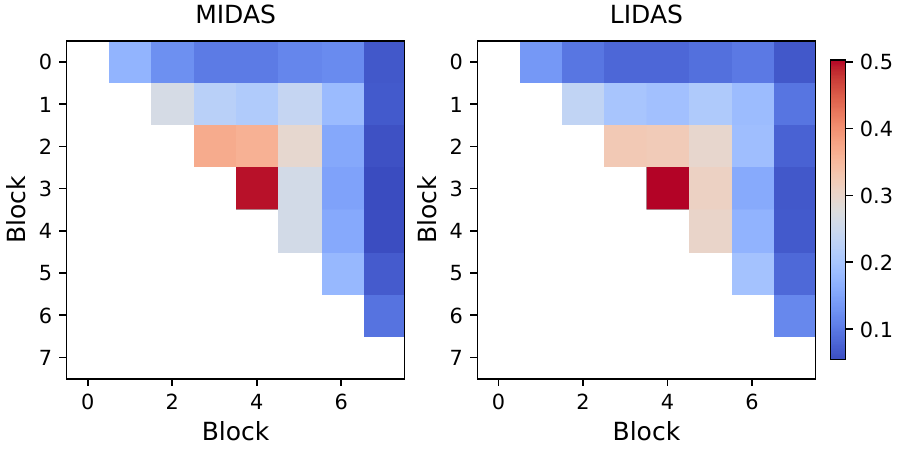}
\caption{Small models (360M): block similarity for \MD and \LD.}
\label{fig:apx:small_block_similarity}
\end{figure*}

For completeness, we also show block-similarity structure at 360M (\cref{fig:apx:small_block_similarity}), which mirrors the symmetry patterns observed at 1.7B (cf. \cref{fig:main_block_similarity}).

\subsection{\changed{Ablating block size}}
\begin{figure*}
    \centering
    \includegraphics[width=\textwidth]{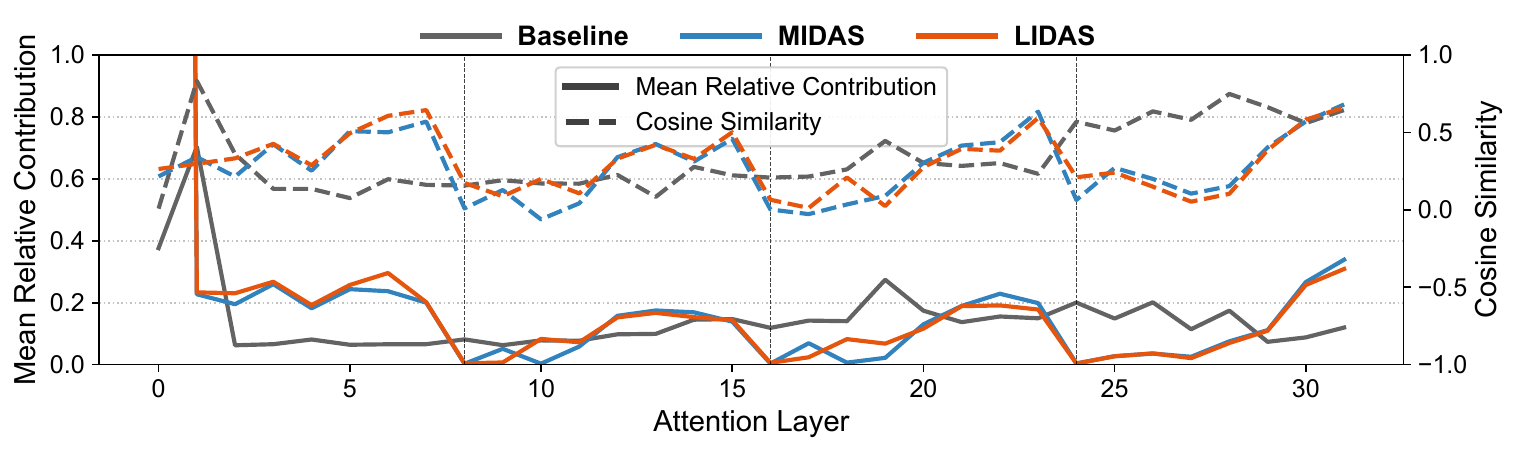}
    \caption{360M models with block size 8: Mean Relative Contribution and Cosine Similarity plots.}
    \label{fig:appx:block_size_8}
\end{figure*}

\begin{figure*}
    \centering
    \includegraphics[width=\textwidth]{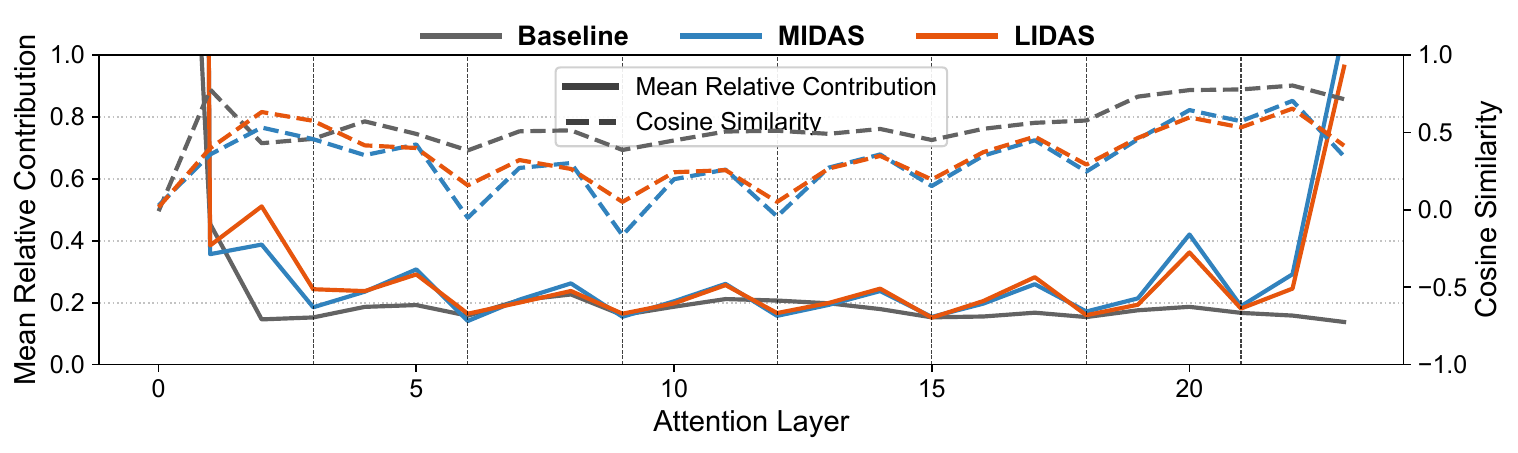}
    \caption{1.7B models with block size 3: Mean Relative Contribution and Cosine Similarity plots.}
    \label{fig:appx:block_size_3}
\end{figure*}

We report mean relative contribution and cosine similarity plots for two ablated model settings: 360M models with block size 8 (\cref{fig:appx:block_size_8}) and 1.7B models with block size 3 (\cref{fig:appx:block_size_3}). As in \cref{fig:main_cyclical_attention}, we observe clear patterns throughout block computation: for the 1.7B \MD and \LD models, both mean relative contribution and cosine similarity peak at the last layer of every block, and in the 360M grown models cosine similarity is likewise maximized at the final layer of each block. These experiments suggest that the emergence of such patterns is a characteristic of the growing training method itself, rather than a peculiarity of a specific block size.

\section{\changed{Layer Norm Scaling}}

LayerNorm-Scaling (\LN \citet{sun2025curse}) is a method that modifies the layer norm sublayer of pre-layernorm transformer architectures with the purpose of increasing the depth usage of later layers. It scales (the variance of) the output $h_l^{\prime}$ of the layer normalization inversely by the square root of its depth
\begin{equation*}
    h_l^{\prime}=\frac{1}{\sqrt{l}}\text{LayerNorm}(h_l)
\end{equation*}

where $h_l$ is the input of the layer norm sublayer.
This simple modification mitigates the output variance explosion of deeper Transformer layers, improving their contribution. Additionally, it preserves the training stability common to all pre-layernorm models, which is demonstrated both theoretically and experimentally.

In our setting, however, \LN does not improve depth utilization according to the three diagnostics we used in \cref{fig:midas:depth_efficiency}. At both scales, the depth score shifts earlier, top-5 early-exit overlap increases at earlier layers, and tuned-lens early-exit accuracy plateaus sooner than for the baseline and grown models (\cref{fig:midas:depth_efficiency,fig:midas:depth_efficiency_360}).

\begin{figure*}[t]
    \centering
  \includegraphics[width=\textwidth]{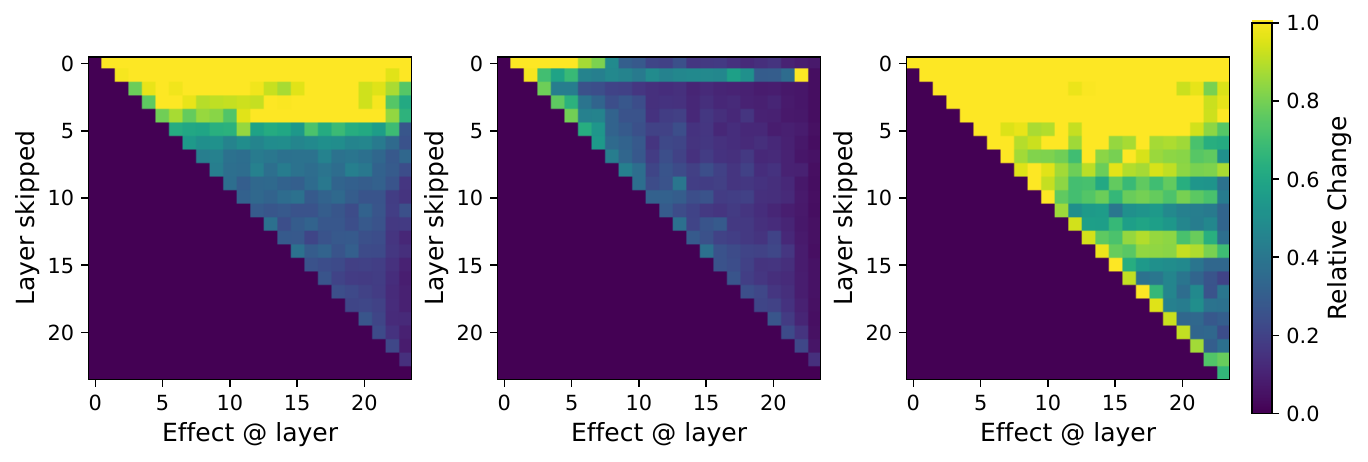}
  \caption{Baseline (1.7 B) model combined with \LN: (from left to right) future propagated layer, future local layer and current attention effects heatmaps}
  \label{fig:apx:ln-scaling-heatmaps-baseline}
\end{figure*}

\begin{figure*}[t]
    \centering
  \includegraphics[width=\textwidth]{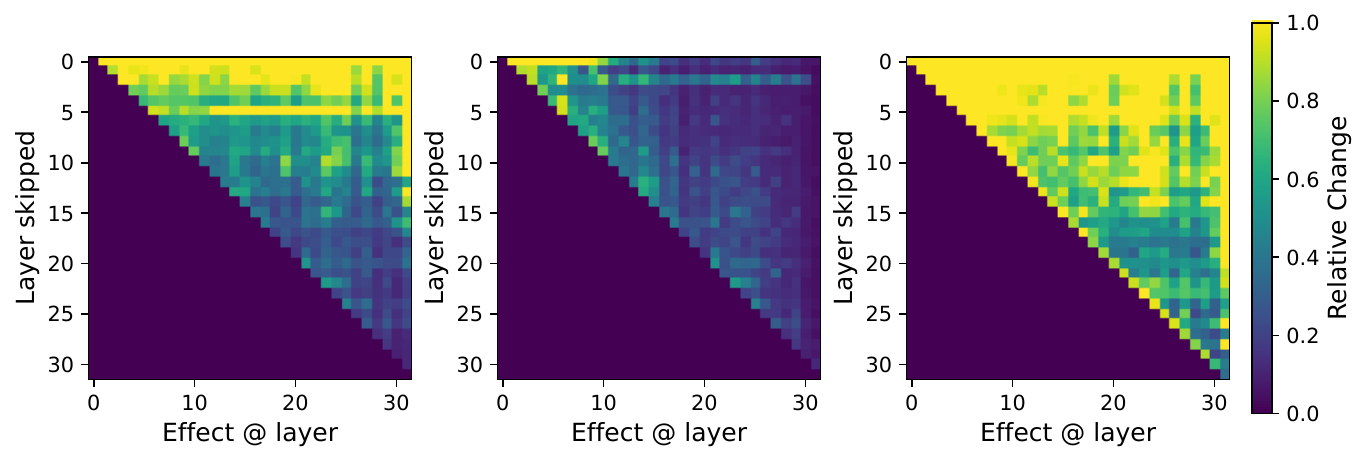}
  \caption{Baseline (360 M) model combined with \LN: (from left to right) future propagated layer, future local layer and current attention effects heatmaps}
  \label{fig:apx:ln-scaling-heatmaps-baseline_360}
\end{figure*}

To probe this further, we evaluate \emph{future-propagated}, \emph{future-local}, and \emph{current-attention} ablations under \LN (see \cref{sec:apx:exp_setup} for definitions of these interventions). Across both 1.7B (\cref{fig:apx:ln-scaling-heatmaps-baseline}) and 360M (\cref{fig:apx:ln-scaling-heatmaps-baseline_360}) models, interventions to later layers produce smaller downstream effects than in the Baseline, \LD, and \MD models, supporting our earlier finding that \LN concentrates computation earlier rather than improving later-layer usage. In line with \cref{tab:exp:results}, the apparent effectiveness of \LN diminishes at larger scales. This scale sensitivity may explain the discrepancy with \citet{sun2025curse}, which does not scale to larger settings.

\begin{figure}[ht]
    \vspace{-5mm}
    \begin{center}
        \includegraphics[width=\textwidth]{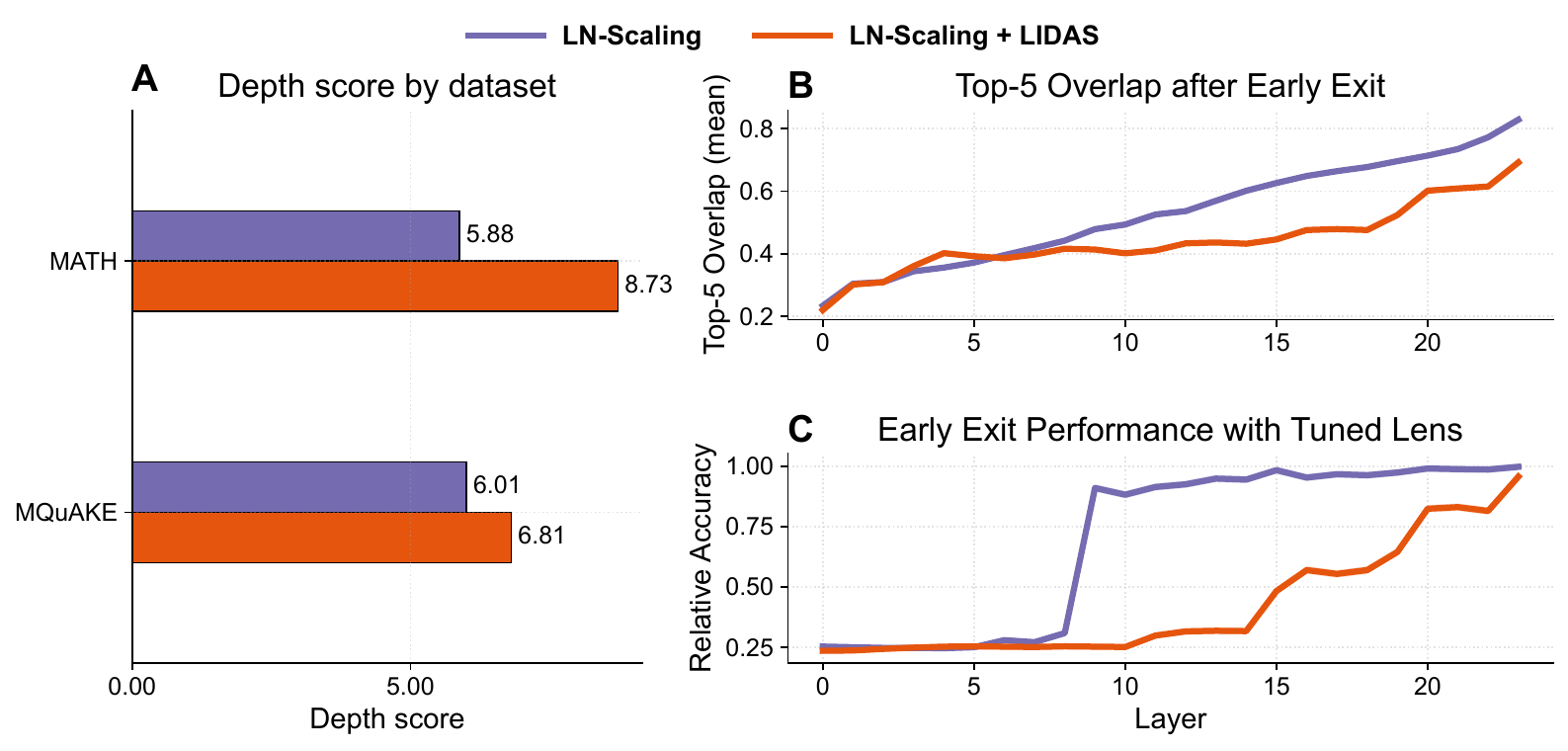}
    \end{center}
    \caption{ \textbf{Depth-grown models use their depth more also when applying growing techniques to \LN (1.7B)}.}
    \label{fig:midas:ln_depth_efficiency_1700}
\end{figure}

\begin{figure}[ht]
    \vspace{-5mm}
    \begin{center}
        \includegraphics[width=\textwidth]{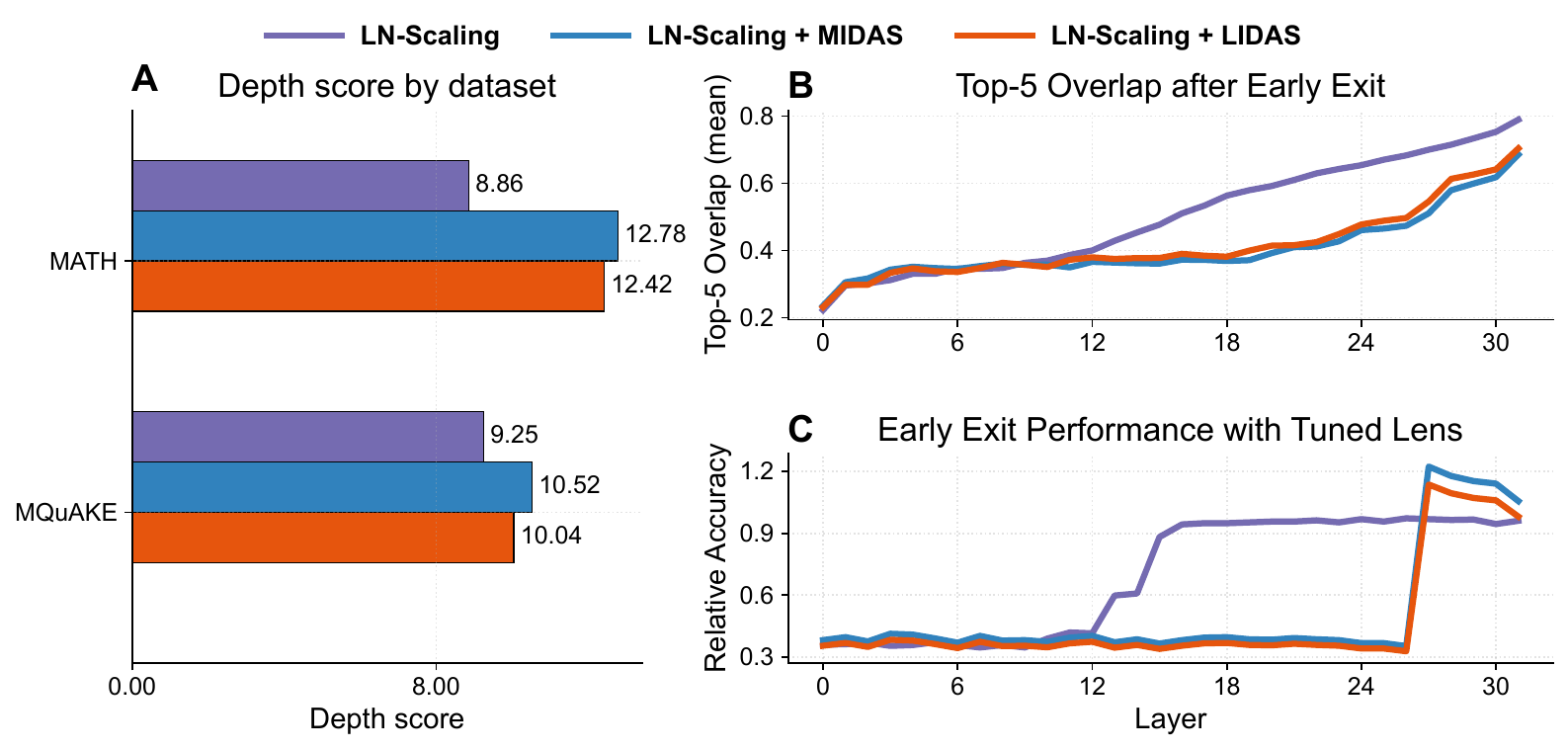}
    \end{center}
    \caption{ \textbf{Depth-grown models use their depth more also when applying growing techniques to \LN (360M)}}
    \label{fig:midas:ln_depth_efficiency_360}
\end{figure}

To understand if \MD and \LD are also effective for this new architecture, we investigated the depth utilization when combining \LN and growing. In \cref{fig:midas:ln_depth_efficiency_1700,fig:midas:ln_depth_efficiency_360} we find, that growing can also increase the depth usage of architectures using \LN, indicating the generality of our findings. 

Furthermore, we find that \LN does not yield substantial gains over the baseline and is typically outperformed by \LD, particularly at the 1.7B model size. When combining \LN with \LD \cref{app:tab:exp_with_ln_scaling:results}, we observe consistent improvements over the LayerNorm-scaled baseline for the 360M model in all categories except Closed-book Q\&A, and in reasoning-heavy tasks also for the 1.7B model.

\begin{figure}[thp!]
    \begin{minipage}[b]{1.0\textwidth}
     \resizebox{0.9999\textwidth}{!}{ 
        \addtolength{\tabcolsep}{-0.2em}
        \begin{tabular}{ll|c|cc|cc|cc|cc}
        \toprule
        \multicolumn{2}{c|}{} &\multicolumn{7}{c|}{Standard cooldown} &\multicolumn{2}{c}{Math cooldown}  \\ \cmidrule(){3-11}
        \multicolumn{2}{c|}{\bf } &\multicolumn{1}{c|}{\bf} &\multicolumn{1}{c}{\bf Open-book} &\multicolumn{1}{c|}{\bf Closed-book}&\multicolumn{1}{c}{\bf} &\multicolumn{1}{c|}{\bf} &\multicolumn{2}{c|}{} &\multicolumn{2}{c}{}  \\
        \multicolumn{2}{c|}{\bf } &\multicolumn{1}{c|}{\bf Holdout Set} &\multicolumn{1}{c}{\bf Q\&A} &\multicolumn{1}{c|}{\bf Q\&A}&\multicolumn{1}{c}{\bf Lambada} &\multicolumn{1}{c|}{\bf Hellaswag } &\multicolumn{1}{c}{\bf Math Word} &\multicolumn{1}{c|}{\bf Primitives } &\multicolumn{1}{c}{\bf Math Word} &\multicolumn{1}{c}{\bf Primitives } \\
        \multicolumn{2}{c|}{\bf } &\multicolumn{1}{c|}{(NLL $\downarrow$)} &\multicolumn{1}{c}{(F1 $\uparrow$)} &\multicolumn{1}{c|}{(F1 $\uparrow$)} &\multicolumn{1}{c}{(Acc $\uparrow$)} &\multicolumn{1}{c|}{(Acc $\uparrow$)} &\multicolumn{1}{c}{(Acc $\uparrow$)} &\multicolumn{1}{c|}{(Acc $\uparrow$)} &\multicolumn{1}{c}{(Acc $\uparrow$)} &\multicolumn{1}{c}{(Acc $\uparrow$)} \\
        \toprule
        \parbox[t]{2mm}{\multirow{4}{*}{\rotatebox[origin=c]{90}{360M}}}
& \LN
& \textbf{2.16}  
& 23.13          
& \textbf{14.89} 
& 42.17          
& 40.00          
& 2.89           
& 31.38          
& 8.45           
& 41.26          
\\
\cmidrule(){2-11}
& \LN + \MD
& 2.19
& 22.04
& 14.05
& 42.77
& 39.84
& \textbf{4.03}  
& 33.04
& 7.72
& 36.90
\\
& \LN + \LD
& \textbf{2.16}  
& \textbf{25.54} 
& 14.12
& \textbf{44.83} 
& \textbf{41.06} 
& 4.00
& \textbf{35.30} 
& \textbf{12.43} 
& \textbf{53.48} 
\\

        \toprule
        \parbox[t]{2mm}{\multirow{2}{*}{\rotatebox[origin=c]{90}{1.7B}}}
& \LN
& 1.97           
& \textbf{29.11} 
& \textbf{18.63} 
& 48.94          
& 45.45          
& 11.00          
& \textbf{44.38} 
& 17.84
& 50.58 
\\
\cmidrule(){2-11}
& \LN + \LD
& \textbf{1.96}  
& 28.04
& 18.42
& \textbf{51.37} 
& \textbf{46.69} 
& \textbf{17.32} 
& 43.32
& \textbf{23.98}
& \textbf{56.28}
\\

        \bottomrule
        \end{tabular}
        }
        \small \captionof{table}{\textbf{Downstream performance of baseline and depth–grown models under \LN.} Compared to \cref{tab:exp:results}, \LD typically improves over the baseline, especially on reasoning-heavy tasks, but the gains are not uniform across datasets, indicating that growth combined with \LN yields architectures with qualitatively distinct behaviour.}
        \label{app:tab:exp_with_ln_scaling:results}
    \end{minipage}
\end{figure}

\end{document}